\documentclass[lettersize,journal]{IEEEtran}

\usepackage[final]{changes} 

\usepackage{algpseudocode}
\usepackage{algorithm}
\usepackage{amsmath}
\usepackage{amssymb}
\usepackage{amsfonts}
\usepackage{array}
\usepackage{cite}
\usepackage{color}
\usepackage{colortbl}
\usepackage{dirtytalk}
\usepackage{graphicx}
\usepackage{multirow}
\usepackage{nowidow}
\usepackage{siunitx}
\usepackage{stfloats}
\usepackage{subcaption}
\usepackage{textcomp}
\usepackage{tikz}
\usepackage{url}
\usepackage{verbatim}
\usepackage{xcolor}
\usetikzlibrary{positioning, shapes.misc}
\usepackage{glossaries}
\usepackage{tabularx}
\usepackage{booktabs}
\usepackage{tablefootnote}
\usepackage{enumitem}
\usetikzlibrary{shapes,backgrounds}
\usetikzlibrary{shapes.misc, positioning}
\usepackage{color, colortbl}
\usepackage{lipsum}     
\usepackage{ntheorem}   
\usepackage{mdframed}   
\usepackage{multicol}
\usepackage{ragged2e}
\usepackage{makecell}
\usepackage{tocloft}

\newglossaryentry{acro:1D}{type=\acronymtype,
	name={1D},
	description={One-dimensional},
	text={1D},
	first={1D},
}
\newglossaryentry{acro:2D}{type=\acronymtype,
	name={2D},
	description={Two-dimensional},
	text={2D},
	first={2D},
}
\newglossaryentry{acro:3D}{type=\acronymtype,
	name={3D},
	description={Three-dimensional},
	text={3D},
	first={3D},
}

\newglossaryentry{acro:AI}{type=\acronymtype,
	name={AI},
	description={Artificial Intelligence},
	text={AI},
    first={Artificial Intelligence~(AI)}, 
}

\newglossaryentry{acro:AMR}{type=\acronymtype,
	name={AMR},
	description={Autonomous Mobile Robot},
	text={AMR},
    first={Autonomous Mobile Robot~(AMR)}, 
    plural={AMR\glspluralsuffix},
    firstplural={Autonomous Mobile Robot\glspluralsuffix\space(AMR\glspluralsuffix)},
}

\newglossaryentry{acro:ANN}{type=\acronymtype,
	name={ANN},
	description={Artificial Neural Network},
	text={ANN},
	first={Artificial Neural Network~(ANN)},
}

\newglossaryentry{acro:API}{type=\acronymtype,
	name={API},
	description={ Application Programming Interface},
	text={API},	
    first={Application Programming Interface~(API)},
    plural={APIs},
	firstplural={Application Programming Interfaces~(APIs)},
}
\newglossaryentry{acro:BIM}{type=\acronymtype,
	name={BIM},
	description={Building Information Modeling},
	text={BIM},
	first={Building Information Modeling~(BIM)},
}

\newglossaryentry{acro:CD}{type=\acronymtype,
	name={CD},
	description={Continuous Delivery},
	text={CD},
	first={Continuous Delivery~(CD)},
}
 \newglossaryentry{acro:CAD}{type=\acronymtype,
	name={CAD},
	description={Computer-Aided Design},
	text={CAD},
	first={Computer-Aided Design~(CAD)},
}
\newglossaryentry{acro:CPU}{type=\acronymtype,
	name={CPU},
	description={Central Processing Unit},
	text={CPU},
	first={Central Processing Unit~(CPU)},	
    firstplural={Central Processing Units~(CPUs)},
    plural={CPUs},
}

\newglossaryentry{acro:CI}{type=\acronymtype,
	name={CI},
	description={Continuous Integration},
	text={CI},
	first={Continuous Integration~(CI)},
}

\newglossaryentry{acro:CV}{type=\acronymtype,
	name={CV},
	description={Computer Vision},
	text={CV},
	first={Computer Vision~(CV)},
}

\newglossaryentry{acro:CNN}{type=\acronymtype,
	name={CNN},
	description={Convolutional Neural Network},
	text={CNN},
	first={Convolutional Neural Network~(CNN)},
        plural={CNNs},
        firstplural={Convolutional Neural Networks~(CNNs)},
}

\newglossaryentry{acro:DL}{type=\acronymtype,
	name={DL},
	description={Deep Learning},
	text={DL},
	first={Deep Learning~(DL)},
}
\newglossaryentry{acro:DoF}{type=\acronymtype,
	name={DoF},
	description={Degree of Freedom},
	text={DoF},
	first={Degree of Freedom~(DoF)},
}

\newglossaryentry{acro:FP}{type=\acronymtype,
	name={FP},
	description={False-Positive},
	text={FP},
	first={False-Positive~(FP)},
}

\newglossaryentry{acro:FPS}{type=\acronymtype,
	name={FPS},
	description={Frames per Second},
	text={FPS},
	first={Frames per Second~(FPS)},
}

\newglossaryentry{acro:FOV}{type=\acronymtype,
	name={FOV},
	description={Field of View},
	text={FOV},
	first={Field of View~(FOV)},
}
\newglossaryentry{acro:GAN}{type=\acronymtype,
	name={GAN},
	description={Generative Adversarial Network},
	text={GAN},
	first={Generative Adversarial Network~(GAN)},
}
\newglossaryentry{acro:GDPR}{type=\acronymtype,
	name={GDPR},
	description={ General Data Protection Regulation},
	text={GDPR},
	first={ General Data Protection Regulation~(GDPR)},
}
\newglossaryentry{acro:GPU}{type=\acronymtype,
	name={GPU},
	description={Graphics Processing Unit},
	text={GPU},
	first={Graphics Processing Unit~(GPU)},
    firstplural={Graphics Processing Units~(GPUs)},
    plural={GPU\glspluralsuffix},
}

\newglossaryentry{acro:HRI}{type=\acronymtype,
	name={HRI},
	description={Human Robot Interaction},
	text={HRI},
	first={Human Robot Interaction~(HRI)},
}

\newglossaryentry{acro:HI}{type=\acronymtype,
	name={HIPer},
	description={Human-Inspired Scene Perception Model},
	text={HIPer-Model},
	first={Human-Inspired Scene Perception~(HIPer)-Model},
}

\newglossaryentry{acro:ICP}{type=\acronymtype,
	name={ICP},
	description={Iterative Closest Point},
	text={ICP},
	first={Iterative Closest Point~(ICP)},
}

\newglossaryentry{acro:IOU}{type=\acronymtype,
	name={IOU},
	description={Intersection over Union},
	text={IOU},
	first={Intersection over Union~(IOU)},
}

\newglossaryentry{acro:ISO}{type=\acronymtype,
	name={ISO},
	description={International Organization for Standardization},
	text={ISO},
	first={International Organization for Standardization~(ISO)},
}

\newglossaryentry{acro:IFR}{type=\acronymtype,
	name={IFR},
	description={International Federation of Robotics},
	text={IFR},
	first={International Federation of Robotics~(IFR)},
}

\newglossaryentry{acro:IMU}{type=\acronymtype,
	name={IMU},
	description={Inertial Measurement Unit},
	text={IMU},
	first={Inertial Measurement Unit~(IMU)},
    plural={IMU\glspluralsuffix},
    firstplural={IMU\glspluralsuffix},
}

\newglossaryentry{acro:JSON}{type=\acronymtype,
	name={JSON},
	description={JavaScript Object Notation},
	text={JSON},
	first={JavaScript Object Notation~(JSON)},
}

\newglossaryentry{acro:KB}{type=\acronymtype,
	name={KB},
	description={Knowledge Base},
	text={KB},
	first={Knowledge Base~(KB)},
}

\newglossaryentry{acro:LiDAR}{type=\acronymtype,
	name={LiDAR},
	description={Light Detection and Ranging},
	text={LiDAR},
	first={Light Detection and Ranging~(LiDAR)},
}
\newglossaryentry{acro:LLM}{type=\acronymtype,
	name={LLM},
	description={Large Language Model},
	text={LLM},
	first={Large Language Model~(LLM)},
        firstplural={Large Language Models~(LLMs)},
        plural={LLMs},
}
\newglossaryentry{acro:LTM}{type=\acronymtype,
	name={LTM},
	description={Long-Term Memory},
	text={LTM},
	first={Long-Term Memory~(LTM)},
}
\newglossaryentry{acro:LSTM}{type=\acronymtype,
	name={LSTM},
	description={Long Short-Term Memory},
	text={LSTM},
	first={Long Short-Term Memory~(LSTM)},
}
\newglossaryentry{acro:MOT}{type=\acronymtype,
	name={MOT},
	description={Multi-Object Tracking},
	text={MOT},
	first={Multi-Object Tracking~(MOT)},
}

\newglossaryentry{acro:MSR}{type=\acronymtype,
	name={MSR},
	description={Multi-Functional Service Robot},
	text={MSR},
	first={Multi-Functional Service Robot~(MSR)},
        firstplural={Multi-Functional Service Robots~(MSRs)},
}

\newglossaryentry{acro:mAP}{type=\acronymtype,
	name={mAP},
	description={Mean Average Precision},
	text={mAP},
	first={mean average precision~(mAP)},
}

\newglossaryentry{acro:NDT}{type=\acronymtype,
	name={NDT},
	description={Normal Distribution},
	text={NDT},
	first={Normal Distribution~(NDT)},
        firstplural={NDT\glspluralsuffix},
        plural={NDT\glspluralsuffix},
}

\newglossaryentry{acro:NLP}{type=\acronymtype,
	name={NLP},
	description={Natural Language Processing},
	text={NLP},
	first={Natural Language Processing~(NLP)},
}

\newglossaryentry{acro:OWL}{type=\acronymtype,
	name={OWL},
	description={Web Ontology Language},
	text={OWL},
	first={Web Ontology Language~(OWL)},
}

\newglossaryentry{acro:RAM}{type=\acronymtype,
	name={RAM},
	description={Random Access Memory},
	text={RAM},
	first={Random Access Memory~(RAM)},
}

\newglossaryentry{acro:ROI}{type=\acronymtype,
	name={ROI},
	description={Region of Interest},
	text={ROI},
	first={Region of Interest~(ROI)},
}
\newglossaryentry{acro:RMS}{type=\acronymtype,
	name={RMS},
	description={Root Mean Square},
	text={RMS},
	first={Root Mean Square~(RMS)},
}
\newglossaryentry{acro:ROS}{type=\acronymtype,
	name={ROS},
	description={Robot Operating System},
	text={ROS},
	first={Robot Operating System~(ROS)},
}

\newglossaryentry{acro:RNN}{type=\acronymtype,
	name={RNN},
	description={Recurrent Neural Network},
	text={RNN},
	first={Recurrent Neural Network~(RNN)},
    firstplural={Recurrent Neural Networks~(RNNs)},
}

\newglossaryentry{acro:RCNN}{type=\acronymtype,
	name={R-CNN},
	description={Region Based Convolutional Neural Network},
	text={R-CNN},
	first={Region Based Convolutional Neural Network~(R-CNN)},
    plural = {R-CNNs},
}

\newglossaryentry{acro:RDBMS}{type=\acronymtype,
	name={RDBMS},
	description={Relational Database Management System},
	text={RDBMS},
	first= {Relational Database Management System~(RDBMS)},       
    firstplural={Relational Database Management System\glspluralsuffix\space(RDBMS\glspluralsuffix)},
}

\newglossaryentry{acro:SFM}{type=\acronymtype,
	name={SFM},
	description={Social Force Model},
	text={SFM},
	first={Social Force Model~(SFM)},
}

\newglossaryentry{acro:SLAM}{type=\acronymtype,
	name={SLAM},
	description={Simultaneous Localization and Mapping},
	text={SLAM},
	first={Simultaneous Localization and Mapping~(SLAM)},
}
\newglossaryentry{acro:SoD}{type=\acronymtype,
	name={SoD},
	description={Service on Demand},
	text={SoD},
	first={Service on Demand~(SoD)},
}

\newglossaryentry{acro:STM}{type=\acronymtype,
	name={STM},
	description={Short-Term Memory},
	text={STM},
	first={Short-Term Memory~(STM)},
}
\newglossaryentry{acro:SQL}{type=\acronymtype,
	name={SQL},
	description={Structured Query Language},
	text={SQL},
	first={Structured Query Language~(SQL)},
}

\newglossaryentry{acro:TP}{type=\acronymtype,
	name={TP},
	description={True-Positive},
	text={TP},
	first={True-Positive~(TP)},
}

\newglossaryentry{acro:ToF}{type=\acronymtype,
	name={TOF},
	description={Time of Flight},
	text={TOF},
	first={Time of Flight~(TOF)},
}
\newglossaryentry{acro:TSDF}{type=\acronymtype,
	name={TSDF},
	description={Truncated Signed Distance Field},
	text={TSDF},
	first={Truncated Signed Distance Field~(TSDF)},
}

\newglossaryentry{acro:UI}{type=\acronymtype,
	name={UI},
	description={User Interface},
	text={UI},
	first={User Interface~(UI)},
}
\newglossaryentry{acro:UUID}{type=\acronymtype,
	name={UUID},
	description={Universally Unique Identifier},
	text={UUID},
	first={Universally Unique Identifier~(UUID)},
}
\newglossaryentry{acro:ViT}{type=\acronymtype,
	name={ViT},
	description={Vision Transformer},
	text={ViT},
	first={Vision Transformer~(ViT)},
        plural={ViTs},
        firstplural={ Vision Transformers~(ViTs)},
}

\usepackage{threeparttable}

\makeglossaries

\begin{document}

\algnewcommand\algorithmicswitch{\textbf{switch}}
\algnewcommand\algorithmiccase{\textbf{case}}
\algnewcommand\algorithmicassert{\texttt{assert}}
\algnewcommand\Assert[1]{\State \algorithmicassert(#1)}%

\algdef{SE}[SWITCH]{Switch}{EndSwitch}[1]{\algorithmicswitch\ #1\ \algorithmicdo}{\algorithmicend\ \algorithmicswitch}%
\algdef{SE}[CASE]{Case}{EndCase}[1]{\algorithmiccase\ #1}{\algorithmicend\ \algorithmiccase}%
\algtext*{EndSwitch}%
\algtext*{EndCase}%

\title{HIPer: A Human-Inspired Scene Perception Model \\ for Multifunctional Mobile Robots}

\author{Florenz Graf$^{1}$, Jochen Lindermayr$^{1}$, Birgit Graf$^{1}$, Werner Kraus$^{1}$, and Marco F. Huber$^{2}$%
\thanks{$^{1}$Department Robot and Assistive Systems, Fraunhofer IPA, 70569 Stuttgart, Germany. {\tt\small<first name>.<last name> @ipa.fraunhofer.de}}%
\thanks{$^{2}$Department Cyber Cognitive Intelligence (CCI), Fraunhofer IPA, and Institute of Industrial Manufacturing and Management IFF, University of Stuttgart, 70569 Stuttgart, Germany. {\tt\small marco.huber@ieee.org}}%
}




\IEEEpubid{\copyright~2024 IEEE \hspace{16cm} }
\maketitle

\begin{abstract}


Taking over arbitrary tasks like humans do with a mobile service robot in open-world settings requires a holistic scene perception for decision-making and high-level control. This paper presents a human-inspired scene perception model to minimize the gap between human and robotic capabilities. The approach takes over fundamental neuroscience concepts, such as a triplet perception split into recognition, knowledge representation, and knowledge interpretation. A recognition system splits the background and foreground to integrate exchangeable image-based object detectors and SLAM, a multi-layer knowledge base represents scene information in a hierarchical structure and offers interfaces for high-level control, and knowledge interpretation methods deploy spatio-temporal scene analysis and perceptual learning for self-adjustment. A single-setting ablation study is used to evaluate the impact of each component on the overall performance for a fetch-and-carry scenario in two simulated and one real-world environment. 

\end{abstract}


\section{Introduction}
\IEEEPARstart{T}{he} vision of a robot that performs daily tasks, such as tidying up, handling dishes, preparing food, and fetching and carrying arbitrary objects, is far away. While these tasks are straightforward for humans, current robots lack the necessary capabilities. One fundamental pillar in open-world settings is scene perception since it allows robots to \emph{understand} and \emph{represent} the world~\cite{Garg2020}. 
Scene perception involves recognizing the environment through semantic knowledge, i.e., perceiving various objects in their spatial and temporal context, which needs continual updating throughout the robot's operational lifespan. Consequently, it forms the knowledge base for cross-task decision-making and high-level control to obtain the required intelligent behavior at high autonomy.

Supporting people is the objective of service robots. Available products for open-world operation take over a distinct service, such as transportation, entertainment, or cleaning~\cite{Mueller2023}. They excel in their designated tasks as their perception is specifically optimized for a particular application rather than a holistic solution. Especially when addressing physical tasks with robots, expensive hardware must be compensated. One trend is to take over multiple services with a single robot~(cp.~\cite{Mueller2023}). Nevertheless, as such multi-functional mobile robots assume control over various tasks, the associated task-specific demands multiply, necessitating a comprehensive, modular, and adaptable solution. A significant hurdle for these versatile robots lies in overcoming the seamless integration of components to attain a holistic scene perception, all the while accommodating the demands of mobile systems—requiring swift setup~\cite{Allahverdi2008}, scalability, and reusability~\cite{Moubarak2012}, real-time reactivity~\cite{Garg2020}, and the capability to flexibly adapt to changes in the environment.
\added{In this context, holistic scene perception refers to the ability of the robot to perceive its scene in terms of recognizing and arranging objects and surfaces in a meaningful way to obtain a high-level understanding}~\cite{dong2023}. \added{A major part of the understanding is obtained through scene knowledge interpretation, i.e., an individual capability to interpret semantics in its context}~(cp. \cite{Garg2020, Graf2022a, dong2023}). 
 
\begin{figure}[!t]
  \centering
  \includegraphics[width=0.95\linewidth]{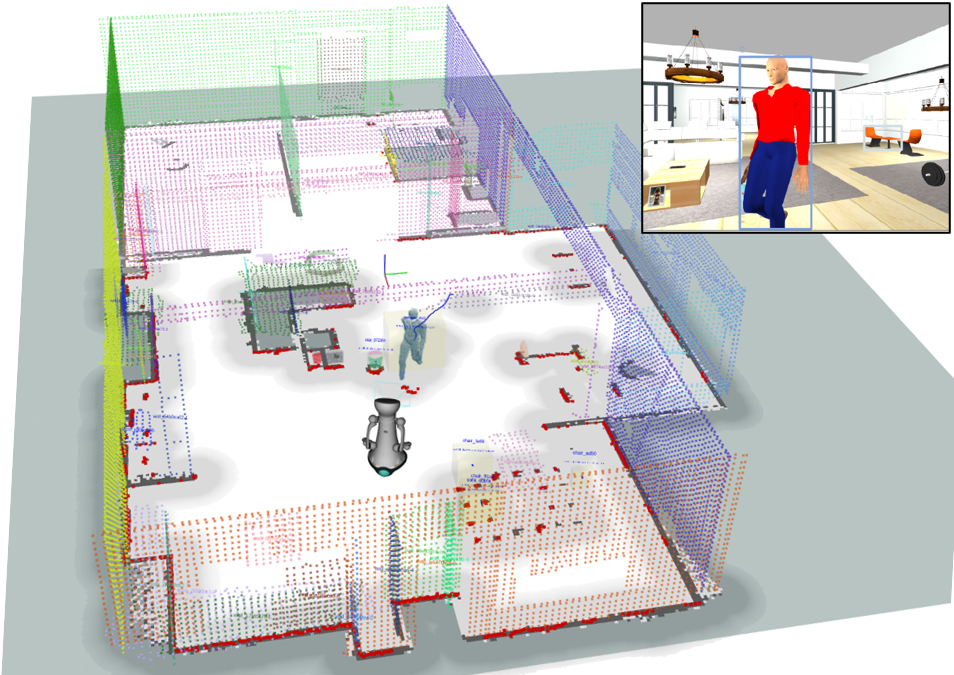}
  \caption{The \gls{acro:HI} provides a holistic scene perception, empowering mobile robots to take over diverse tasks in open-world settings.}
  \label{figures:hiper}
\end{figure}

Recognizing the limited research for such \added{holistic scene perception} solutions, we propose a biologically-inspired approach. Since humans possess comprehensive perceptual capabilities that allow them to manage multiple arbitrary tasks effortlessly, we hypothesize that merging neuroscience concepts of human perception with robotic techniques could bridge the gap between human and robot perception and simplify their collaboration as a side effect. Thus, we developed \gls{acro:HI}~(cp. Fig.~\ref{figures:hiper}) to empower mobile robots through a human-like problem understanding for taking over arbitrary tasks within the open world. 
The \gls{acro:HI} builds upon our recently published paper~\cite{Graf2022a} that transfers the latest studies and theories of the human perceptual process to mobile robots. The transfer points out numerous similarities between theories of human perception that date back many years and comparatively recent research on robot perception. It highlights the limitations of today's robotic perception while showing how humans manage to obtain a holistic perception required for arbitrary tasks. Humans learn an individual scene perception, relying foremost on visual information, throughout their lives. However, initial perception skills are essential for rapid deployment.

This paper considers these identified challenges by providing three substantial contributions:
\begin{itemize}
    \item A comprehensive analysis of the transferability of human perception concepts to mobile robots for designing scene perception models while considering the limitations and potentials of robots. 
    \item The \gls{acro:HI}, a technical solution that merges neuroscience concepts with robot perception to bridge the two most significant shortcomings: providing a holistic scene understanding for cross-task applications and \added{semantic scene} knowledge interpretation using long-term scene analysis.  
    \item Introduction of a novel ablation-study-based benchmark for robotic scene models to fill the gap in research for analyzing their long-term performance. It is deployed in real and simulated environments to demonstrate the perceptual and task performance of the \gls{acro:HI}.
    \end{itemize}

The subsequent sections of this paper follow this structure: Section~\ref{sec:rw} presents relevant literature on specific scene perception techniques and their integrated solutions for mobile robots. Section~\ref{sec:concept} formulates the problem and introduces how the \gls{acro:HI} aims to solve it by presenting its concept and a system overview. Sections~\ref{sec:reco} to~\ref{sec:inter} elaborate on the methodology of each component, which is divided into information recognition, knowledge representation, and interpretation—the three major parts of the \gls{acro:HI}. Section~\ref{sec:eval} details experiments, including a single-setting ablation study in one real and two simulated environments. Finally, this paper concludes with discussions in Section~\ref{sec:discuss} and explores possible future research directions in Section~\ref{sec:conclude}.

\section{Related Work}
\label{sec:rw}
\added{Research in robot scene perception encompasses diverse fields that aim to empower the robot for specific tasks through sensor data processing.} The primary interest lies in \added{visual} perception techniques, which build upon Computer Vision methods. They reach many years into the past and have been adapted to address specific challenges in the context of mobile robots, such as visual \gls{acro:SLAM}, object recognition, semantic segmentation and \gls{acro:MOT}. Recently, research started merging several techniques to achieve a more comprehensive understanding of the scene, \added{which is particularly important for the open-world context}.

\subsubsection*{SLAM} It is a vast research field in robotics for metric scene reconstruction~\cite{Pintore2020}. The robot acquires a spatial environment map through sensor data registration while simultaneously localizing itself to this map~\cite{Thrun2005}. Traditional approaches use 2D-LiDAR data for scan-matching to generate a 2D map, while newer visual SLAM utilizes RGB-D or stereo cameras to generate a 3D map~\cite{Khoyani2023}. The latest approaches rely on multiple sensors for precise localization. For instance, they use dead reckoning with IMU, ICP based on LiDAR, wheel odometry, and visual odometry to minimize ego-motion drift~\cite{Labbe2018}. Localization techniques include Bayes Filter-based algorithms (Particle Filter, Extended Kalman Filter) and optimization algorithms like graphs or bundle adjustment. In contrast to so-called \emph{full-SLAM}, \emph{online SLAM} solely recovers the recent poses, allowing a constant memory allocation~\cite{Labbe2018, MurArtal2017}. Moreover, modern approaches feature multi-robot mapping~\cite{Tian2022}, WIFI signal mapping~\cite{Labbe2018}, or long-term map updates~\cite{Doerr2020}, which are crucial for dynamic environments. However, since SLAM is restricted to spatial data, supplementary recognition techniques for clustering regions and assigning the semantics are necessary for high-level scene understanding~\cite{Kim2020}.

\subsubsection*{Object recognition} Mobile robots benefit significantly from two key aspects of visual recognition: object detection and semantic segmentation. These fields both involve the analysis of images or point clouds at a specific moment, but they differ in the representation of their output. 

The prevailing approach in image-based object detection relies on \gls{acro:DL} models like \glspl{acro:CNN}, such as YOLO~\cite{Redmon2016}, InternImage~\cite{Wang2022}, or \glspl{acro:ViT}, such as SWIN~\cite{Liu2022}, ConvNet~\cite{Liu2022a}, and DETR~\cite{Carion2020}. When trained on extensive datasets, these methods achieve an impressive \gls{acro:mAP} of nearly \SI{65}{\percent}. While CNN-based detectors offer a robust initial estimation, their deployment on robots demands additional techniques to enhance reliability. Specialized networks also exist for tasks like human skeleton detection, providing joint information~\cite{Nguyen2016, Cao2019}, or (re)detecting specific places~\cite{Lowry2016}. Unsupervised approaches, such as zero-shot detectors like CLIP~\cite{Radford2021}, have gained popularity for their ease of deployment across diverse objects. However, they cannot localize objects, making them only conditionally applicable for mobile robots. 
Although object detectors localize instances within a bounding box, their additional techniques become essential for effectively matching the corresponding 3D data. Vice versa, detectors based on 3D data face challenges in real-time performance and large-scale datasets due to data volume and availability. As a result, such approaches have yet to be popular on mobile robots. 

A practical solution to address the computational demands involves mapping 2D detections into 3D space through semantic segmentation. This method, thoroughly explored in studies~\cite{Kostavelis2015, Garg2020}, assigns semantic labels to each pixel in an image, offering information on object boundaries that can be projected into 3D space. Instance segmentation, a subset of semantic segmentation, provides distinct object masks, with methods like Mask R-CNN~\cite{He2017} being popular. Semantic segmentation proves beneficial for mobile robots where tasks demand precise object localization and tracking. However, generating datasets for segmentation is more time-intensive than bounding-box-based approaches, requiring pixel-wise labeling. Consequently, the available pool of datasets for semantic segmentation is smaller, an important factor when dealing with less common objects~\cite{Kostavelis2015, Premebida2019, Poux2019a}.

\subsubsection*{\gls{acro:MOT}} It involves converting time-discrete object detections into continuous tracks, a crucial aspect for monitoring objects over time. Typical \gls{acro:MOT} integrates Kalman Filters or Particle Filters with \gls{acro:2D} or \gls{acro:3D} object detections that can be split into generative, discriminative, and hybrid approaches~\cite{Chen2019b}. Generative trackers model the entire object appearance, generating an appearance model and matching it to subsequent frames based on metrics like \gls{acro:IOU}. Discriminative trackers learn object-background boundaries using CNN, ViT, or LSTM~\cite{Zhang2020, Wojke2017, Chu2023, Kim2021}. Equivalent to object recognition, \gls{acro:3D} data are preferred over \gls{acro:2D} for mobile robots. Popular approaches use a Hungarian algorithm to process detections before Kalman Filters track objects in \gls{acro:3D} space~\cite{Weng2020, Pang2023, Mueller2021}. Despite addressing ego-motion, sensor and \gls{acro:FOV} limitations impact object perception, necessitating consideration of uncertainties, occlusions, and re-appearances.

\subsubsection*{Integrated Approaches} Combining multiple techniques within a unified framework becomes crucial for a comprehensive scene perception, allowing the recognition of distinct semantics from various places and times. Semantic mapping, also known as metric-semantic reconstruction, has become an upcoming trend in this context, which merges recognition with \gls{acro:SLAM}~\cite{Rosinol2021}. In particular, it enriches spatial environment data from SLAM with semantic scene information, such as generated from semantic segmentation~\cite{Minaee2021} to address the challenge of achieving an abstract and descriptive representation of the entire scene~\cite{Kim2020}. Examples are provided in: \cite{SalasMoreno2013, Mccormac2018, Poux2019}. \added{A notable application, as demonstrated by Langer et al.}~\cite{Langer2022}, \added{involves clustering objects on planes to organize common tasks. Additionally, extracting scene backgrounds by planes contributes to understanding the environmental layout as demonstrated in}~\cite{Chen2019, Suchan2017, Ambrus2017}. 

\added{Semantic mapping has a drawback in its failure to consider object dynamics, as the generated map merely reflects the state of the initial reconstruction. Given the presence of expected moving objects, addressing motion becomes crucial. Recent advancements offer innovative strategies for managing various object states, including static, moving, removed, and new ones. These strategies facilitate instance-specific handling, applicable in diverse scenarios. For example, temporal observation of people's skeletons enables reasoning about actions involving objects}~\cite{Lee2020}. \added{The complexity escalates due to the incorporation of pose optimization for robot localization combined with scene dynamics. To address this, scene graphs have been introduced to anchor objects and humans in hierarchical topological structures}~\cite{Chen2019, Rosinol2021, Hughes2022}. \added{For instance, the \emph{Kimera} framework}~\cite{Rosinol2021} \added{employs scene graphs for multi-level instance-specific reconstruction. This approach allows novel strategies, such as filtering people during mapping, proving particularly beneficial in crowded areas.}

A more intricate approach to handling the acquired scene knowledge is \emph{Knowrob}~\cite{Beetz2018}, a framework for knowledge representation and reasoning. \emph{Knowrob} can describe a complex scene by linking different entities through elaborate ontologies and managing them with various memory concepts, such as episodic or semantic memory. \added{In this context, open-vocabulary mapping has emerged as an important research field, which holds significant relevance for open-world scene understanding since it is not confined to a predefined set of semantic classes. Recent approaches, such as those presented in}~\cite{huang2023, shafiullah2022, gadre2023}, \added{incorporate open-vocabulary properties into scene mapping and its representation. This integration helps to accommodate a broader range of vocabularies and environmental complexities. Such contributions pave the way for more robust and adaptable scene representations, facilitating improved navigation and understanding within diverse environments.}

\added{An alternative strategy to handle open-world scenes is to employ learning strategies.} Unlike approaches that train their perception before real-world deployment, there is scarce research on perceptual lifelong learning. Examples in the early stages are presented in~\cite{Wyatt2011, Dong2022}, \added{aiming to adapt the perception during acting in the real world.} However, the considerable potential of this research field has yet to be integrated.
\section{Concept and System Overview}
\label{sec:concept}

The considered problem of scene perception can be formulated mathematically as
\begin{align}
        X=\sum_{t=0}^\infty x_{t}\quad \xrightarrow{\text{Scene Model}} \quad Y=\sum_{s=0}^k y_{s}~,
    \label{eq:problem}
\end{align}
where a time-continuous raw sensor input $X$ has to be mapped from the scene model to time-discrete understandable (in terms of semantics) scene information $Y$. The sensor input~$X$ consists of continuously captured scene information~$x$ from one or multiple sensors at a given time~$t$. The high-level control system obtains specific scene information $y_s$, which is provided by demand~$k$ based on their needs. The scene perception of the robot, i.e., the scene model, is responsible for handling the transformation from $X$ to $Y$, a prerequisite to operating and taking over multiple and complex tasks.

Although robotic research addresses this complex issue with integrated approaches, combining multiple sub-components working seamlessly, they do not empower mobile robots with sophisticated scene knowledge to take over arbitrary tasks in open-world settings as humans do. When studying neuroscience concepts and theories of human perception and comparing them with robotic approaches, as presented in our previous work, a notable deficiency refers to the scope of perception, i.e., holistic solutions that cover the entire perception process similar to humans~\cite{Graf2022a}. Moreover, related approaches are not designed to extend or replace components, making deploying them to a domain different from the original tasks challenging. One reason is that they do not follow a standardized structure since their architecture is optimized for individual tasks. 

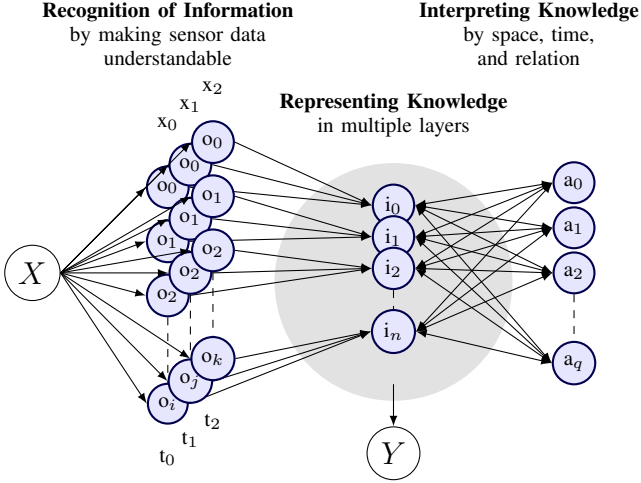
\begin{figure}[tb]
    \centering
    \tikzset{lnode/.style = {
        circle, 
        draw=blue!30!black, 
        thick,
        fill=blue!10,
        inner sep=2.3pt,
        minimum size=15pt }}
    \tikzset{uedge/.style = {
        draw=blue!20!black, 
        very thick}}
    
    \begin{tikzpicture}[scale=0.60]
    \footnotesize
     
    \node[align=center] at (-5.0,4.80) {\textbf{Recognition of Information} \\ by making sensor data \\ understandable};
    \node[align=center] at (0.0,3.0) {\textbf{Representing Knowledge} \\ in multiple layers};
    \node[align=center] at (3.0,4.80) {\textbf{Interpreting Knowledge} \\ by space, time, \\ and relation};
     
    \fill[gray,opacity=0.23] (0,-0.7) circle (75pt);

    \node  (x0)  at (-5, 2.8) {x$_{0}$};
    \node  (x1)  at (-4.5, 3.2) {x$_{1}$};
    \node  (x2)  at (-4, 3.6) {x$_{2}$};
    \node  (t0)  at (-5, -4.6) {t$_{0}$};
    \node  (t1)  at (-4.5, -4.2) {t$_{1}$};
    \node  (t2)  at (-4, -3.8) {t$_{2}$};

    \node[lnode]  (e1)  at (0, 1) {i$_{0}$};
    \node[lnode]  (e2)  at (0, 0.3) {i$_{1}$};
    \node[lnode]  (e3)  at (0, -0.4) {i$_{2}$};
    \node[lnode]  (ep)  at (0, -1.8) {i$_{n}$};
    
    \node[draw,circle,align=center]  (X1)  at (-8, -0.5) {\large $X$};
    \node[lnode]  (o11)  at (-5, 1.4) {o$_{0}$};
    \node[lnode]  (o21)  at (-5, 0.2) {o$_{1}$};
    \node[lnode]  (o31)  at (-5, -1.0) {o$_{2}$};
    \node[lnode]  (oj1)  at (-5, -3.4) {o$_{i}$};
    \draw[dashed] (o31.south) -- (oj1.north);

    \draw[-latex] (o11.east) -- (e1.west);
    \draw[-latex] (o21.east) -- (e2.west);
    \draw[-latex] (o31.east) -- (e3.west);
    \draw[-latex] (oj1.east) -- (ep.west);

    \node[lnode]  (o12)  at (-4.5, 1.9) {o$_{0}$};
    \node[lnode]  (o22)  at (-4.5, 0.7) {o$_{1}$};
    \node[lnode]  (o32)  at (-4.5, -0.5) {o$_{2}$};
    \node[lnode]  (oj2)  at (-4.5, -2.9) {o$_{j}$};
    \draw[dashed] (o32.south) -- (oj2.north);

    \draw[-latex] (o12.east) -- (e1.west);
    \draw[-latex] (o22.east) -- (e2.west);
    \draw[-latex] (o32.east) -- (e3.west);
    \draw[-latex] (oj2.east) -- (ep.west);

    \node[lnode]  (o13)  at (-4, 2.4) {o$_{0}$};
    \node[lnode]  (o23)  at (-4, 1.2) {o$_{1}$};
    \node[lnode]  (o33)  at (-4, 0) {o$_{2}$};
    \node[lnode]  (oj3)  at (-4, -2.4) {o$_{k}$};
    \draw[dashed] (o33.south) -- (oj3.north);

    \draw[-latex] (o13.east) -- (e1.west);
    \draw[-latex] (o23.east) -- (e2.west);
    \draw[-latex] (o33.east) -- (e3.west);
    \draw[-latex] (oj3.east) -- (ep.west);

    \draw[-latex] (X1.east) -- (o11.west);
    \draw[-latex] (X1.east) -- (o21.west);
    \draw[-latex] (X1.east) -- (o31.west);
    \draw[-latex] (X1.east) -- (oj1.west);

    \draw[-latex] (X1.east) -- (o12.west);
    \draw[-latex] (X1.east) -- (o22.west);
    \draw[-latex] (X1.east) -- (o32.west);
    \draw[-latex] (X1.east) -- (oj2.west);

    \draw[-latex] (X1.east) -- (o13.west);
    \draw[-latex] (X1.east) -- (o23.west);
    \draw[-latex] (X1.east) -- (o33.west);
    \draw[-latex] (X1.east) -- (oj3.west);

    \draw[dashed] (e3.south) -- (ep.north);

    \node[lnode]  (a0)  at (4, 1.5) {a$_{0}$};
    \node[lnode]  (a1)  at (4, 0.5) {a$_{1}$};
    \node[lnode]  (a2)  at (4, -0.5) {a$_{2}$};
    \node[lnode]  (aq)  at (4, -2.5) {a$_{q}$};

    \draw[latex-latex] (e1.east) -- (a0.west);
    \draw[latex-latex] (e2.east) -- (a0.west);
    \draw[latex-latex] (e3.east) -- (a0.west);
    \draw[latex-latex] (ep.east) -- (a0.west);
    
    \draw[latex-latex] (e1.east) -- (a1.west);
    \draw[latex-latex] (e2.east) -- (a1.west);
    \draw[latex-latex] (e3.east) -- (a1.west);
    \draw[latex-latex] (ep.east) -- (a1.west);
    
    \draw[latex-latex] (e1.east) -- (a2.west);
    \draw[latex-latex] (e2.east) -- (a2.west);
    \draw[latex-latex] (e3.east) -- (a2.west);
    \draw[latex-latex] (ep.east) -- (a2.west);
    
    \draw[latex-latex] (e1.east) -- (aq.west);
    \draw[latex-latex] (e2.east) -- (aq.west);
    \draw[latex-latex] (e3.east) -- (aq.west);
    \draw[latex-latex] (ep.east) -- (aq.west);
    \draw[dashed] (a2.south) -- (aq.north);
    
    \node[align=center]  (DBS)  at (0 ,-2.8) {};
    \node[draw,circle,align=center]  (Y)  at (0 ,-4.5) {\large $Y$};
    \draw[-latex] (DBS.south) -- (Y.north);

    \end{tikzpicture}
\caption{HIPer model design inspired by human perception consisting of a triplet split of observations~$o$, based on sensory input~$X$, its aggregation into instances~$i$, and spatio-temporal analyses~$a$. The obtained scene knowledge $Y$ is accessible for decision-making and planning.}
\label{figures:overview}
\end{figure}

The \gls{acro:HI} addresses these limitations by merging human perception concepts with the latest robotics approaches. Although human perception is an interconnected formation of many pieces, the three-part division of perception which is presented in common neuroscience theories~\cite[Chapter~3]{Ward2015} allows a clear structure: (I)~recognition makes the sensory input understandable, (II)~different memory types store and represent knowledge in multiple layers, and~(III) interpretation analyzes available knowledge by space, time, and relation.
While related approaches concentrate foremost on sensor-close recognition tasks, as it is the origin of knowledge generation~(cp.~\cite{Neisser1993}), the \gls{acro:HI} aims to provide a holistic solution of the entire perception process to transform the input~$X$ to the output~$Y$ by combining both domains' advantages. In particular, its concept mimics neuroscience's three-part division, offering a clear description of functionalities, as depicted by its design in Fig.~\ref{figures:overview}.

The recognition system aims to make simultaneously captured sensor data understandable by obtaining scene observations~$o$. The \gls{acro:HI} combines visual recognition techniques to extract understandable pieces of environment information, denoted as observations $o$, based on visual sensor data at time~$t$. In particular, the approach processes the foreground and background differently to leverage their characteristics.
The split imitates human's preattentive and postattentive processing~\cite{Wolfe2000}, i.e., the background, characterized through large data with little richness of information, is extracted during the initial reconstruction, allowing for high attention on particular instances in the foreground. The accumulation of all observations~$O$ describes the union of foreground and background observations~$O_b(t)$, and $O_f(t)$, respectively, recognized at time~$t$. It is denoted as
\begin{equation}
O = \bigcup_{t=0}^{\infty}  (O_b(t) \cup O_f(t)) \quad \text{.}
    \label{eq:O}
\end{equation}
Both terms describe a set of $b$, respectively $f$ observations, according to
\begin{equation}
    O_b(t) = \bigcup_{b} o_b \quad \text{and} \quad O_f(t) = \bigcup_{f} o_f \quad \text{.}
    \label{eq:Obf}
\end{equation}
Multiple corresponding scene observations are fused by time and space to a so-called instance~$i$ according to
\begin{equation}
    I = \bigcup_{j=0}^n i_{j} \text{ with } i_j = \bigcup_{l} o_{l} \quad \text{.}
    \label{eq:I}
\end{equation}
The concept of foreground or background instances is equivalent. They are generated from matched observations~$o_f$. While the background instances consider only the robot poses, the foreground instances are generated through a \gls{acro:MOT} pipeline to consider the additional dynamics of the particular instance. 

The obtained scene information, i.e., the sum of observations~$O$ and instances~$I$, are organized and stored in a knowledge base. While observations are anchored to the sensor origin, instances are transformed into the global map frame to obtain a virtual representation of the world. This procedure allows post-processing instances after pose optimizations by recalculating the merging of each instance. Furthermore, the knowledge base~$K$ manages scene analyses~$A$, resulting in 
\begin{equation}
    K = O \bigcup I \bigcup A \quad \text{.}
    \label{eq:E}
\end{equation}
Knowledge interpretation builds upon the preceding steps—knowledge acquisition being a prerequisite—while it is only marginally addressed in related work. The \gls{acro:HI} incorporates this crucial element through the implementation of a union~$p$ of spatio-temporal scene analysis techniques~$a_m$, denoted as~$A$ according to
\begin{equation}
A = \bigcup_{m=0}^p a_{m} \quad \text{.}
    \label{eq:A}
\end{equation}
While robots use pre-trained recognition on extensive datasets for quick deployment, they lack adaptability when deployed to a random environment. 
The analysis techniques aim to observe the particular environment over a longer period to improve scene understanding and reshape pre-trained recognition capabilities during long-term operations.

\begin{figure*}[!ht]
    \centering

    \resizebox{\textwidth}{!}
    {
        \tikzset{lnode/.style = {
        rectangle, 
        align=center,
        fill=blue!10,
        minimum width=1cm}}

        \tikzset{fnode/.style = {
        draw, 
        lightgray,
        dotted,
        align=center,
        fill=white,
        minimum height=5cm, 
        line width=0.7mm}}

        \tikzset{gnode/.style = {
        rectangle, 
        align=center,
        fill=gray!10,
        minimum width=1cm}}

       \begin{tikzpicture}[node distance = 2cm, auto]

         \node[fnode, minimum width=7.6cm] at (-4.8,0.2) {};
         \node[fnode, minimum width=4cm] at (1,0.2) {};
         \node[fnode, minimum width=3.3cm] at (4.6, 0.2) {};

         \small
         \node[align=center] at (-9.8,2.8) {\textbf{Sensory Input} \\ \textbf{$X$}};
         \node[align=center] at (-4.8,3) {\textbf{Recognition of Information}};
         \node[align=center] at (1,3) {\textbf{Knowledge Base}};
         \node[align=center] at (4.6,3) {\textbf{Knowledge Interpretation}};
         \node[align=center] at (1,-2.8)(y) {\textbf{Scene Information Y}};

         \draw[dotted, white, fill=blue!10]
         (1,2.1) coordinate (B) -- 
         (3,-1.9) coordinate(A)-- 
         (-1,-1.9) coordinate(C) --
         cycle;

         \draw[dotted, white, fill=blue!15]
         (1,2.1) coordinate (B) -- 
         (2.6,-1.1) coordinate(A)-- 
         (-0.6,-1.1) coordinate(C) --
         cycle;
         
         \draw[dotted, white, fill=blue!10]
         (1,2.1) coordinate (B) -- 
         (2.2,-0.3) coordinate(A)-- 
         (-0.2,-0.3) coordinate(C) --
         cycle;
             
         \draw[dotted, white, fill=blue!15]
         (1,2.1) coordinate (B) -- 
         (1.8,0.5) coordinate(A)-- 
         (0.2,0.5) coordinate(C) --
         cycle;

         \node at (-9.8, 0.0)(cam) { \includegraphics[width=0.13\linewidth,scale=1]{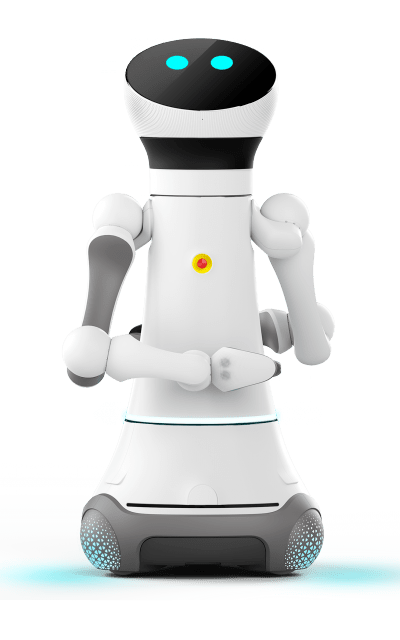}};

         \scriptsize
         \node[align=center, fill=white!10,opacity=.6, text opacity=1] at (-9.8 , -0.2)(sensors) {RGB-D \\ LiDAR \\ Odometry};

         \footnotesize
         \node at (-1.0,2.0)(obs_img1){\includegraphics[height=0.075\linewidth]{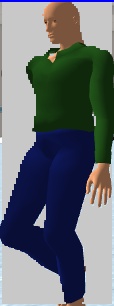}}; 
         \node at (-0.5,2.0)(obs_img2){\includegraphics[height=0.075\linewidth,scale=1]{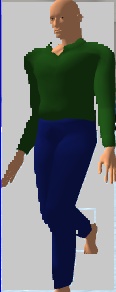}}; 
         \node at (0,2.0)(obs_img3){\includegraphics[height=0.075\linewidth,scale=1]{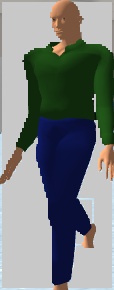}}; 


         \node[gnode] at (-7.9,2)(cnn) {Object \\ Detector};    
         \node at (-6.5,2)(cnn_img){\includegraphics[width=0.09\linewidth, height=0.065\linewidth]{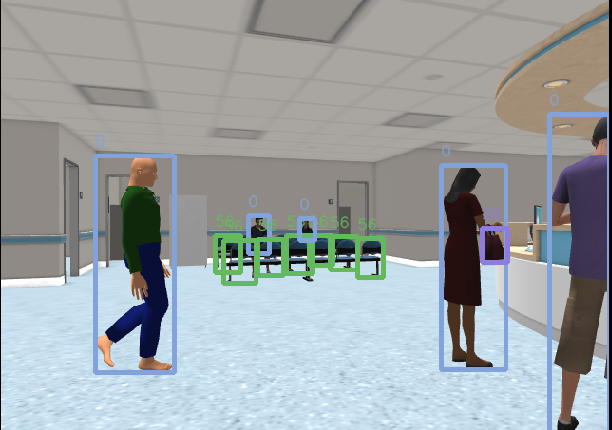}}; 

         \path [draw, -latex] (sensors.north) |- node[near end]{RGB}(cnn);
         \node[gnode,minimum width=1.8cm] at (-7.6,-2.0)(slam) {SLAM};       
         \path [draw, -latex] (sensors.south) |- node[near end]{All}(slam);
        
         \node[lnode,minimum width=1.8cm,minimum height=0.8cm] at (-7.6,0.5)(segF) {Foreground \\ Segmentation};
         \node[lnode,minimum width=1.8cm,minimum height=0.8cm] at (-7.6,-0.7)(segB) {Background \\ Segmentation};
         \node at (-5.9,0.7)(segF_img){\includegraphics[width=0.09\linewidth, height=0.065\linewidth]{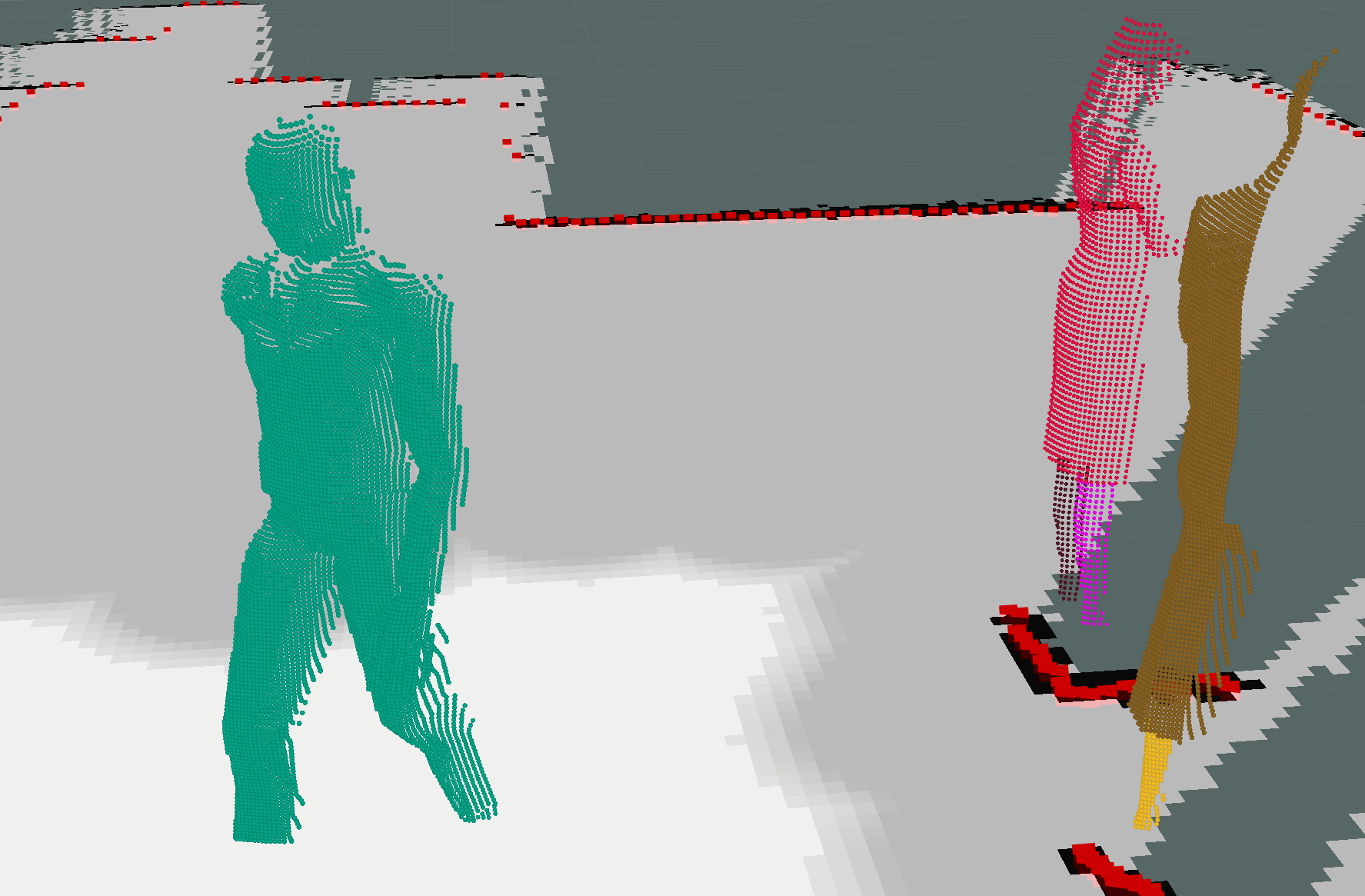}}; 
         \path [draw, -latex, dotted] (segB) -- (segF);
         \path [draw, -latex] (slam.north) -| node[near end]{Points}(segB);

         \path [draw, -latex] (sensors.north) |- node[near end]{Points}(segF);  
         \node[lnode] at (-3.5,2.0)(matchF) {Match};    
         \path [draw, -latex] (cnn_img) -- node{Detection 2D}(matchF);
         \path [draw, -latex] (segF_img) -| node[near start]{Blob 3D}(matchF.west);

         \node[lnode,minimum width=1.3cm,minimum height=0.8cm] at (-3,0.5)(track) {MOT};    
         \path [draw, -latex] (matchF) -| (track);
         \node at (-2,0.9)(obj_img){\includegraphics[width=0.055\linewidth,scale=1]{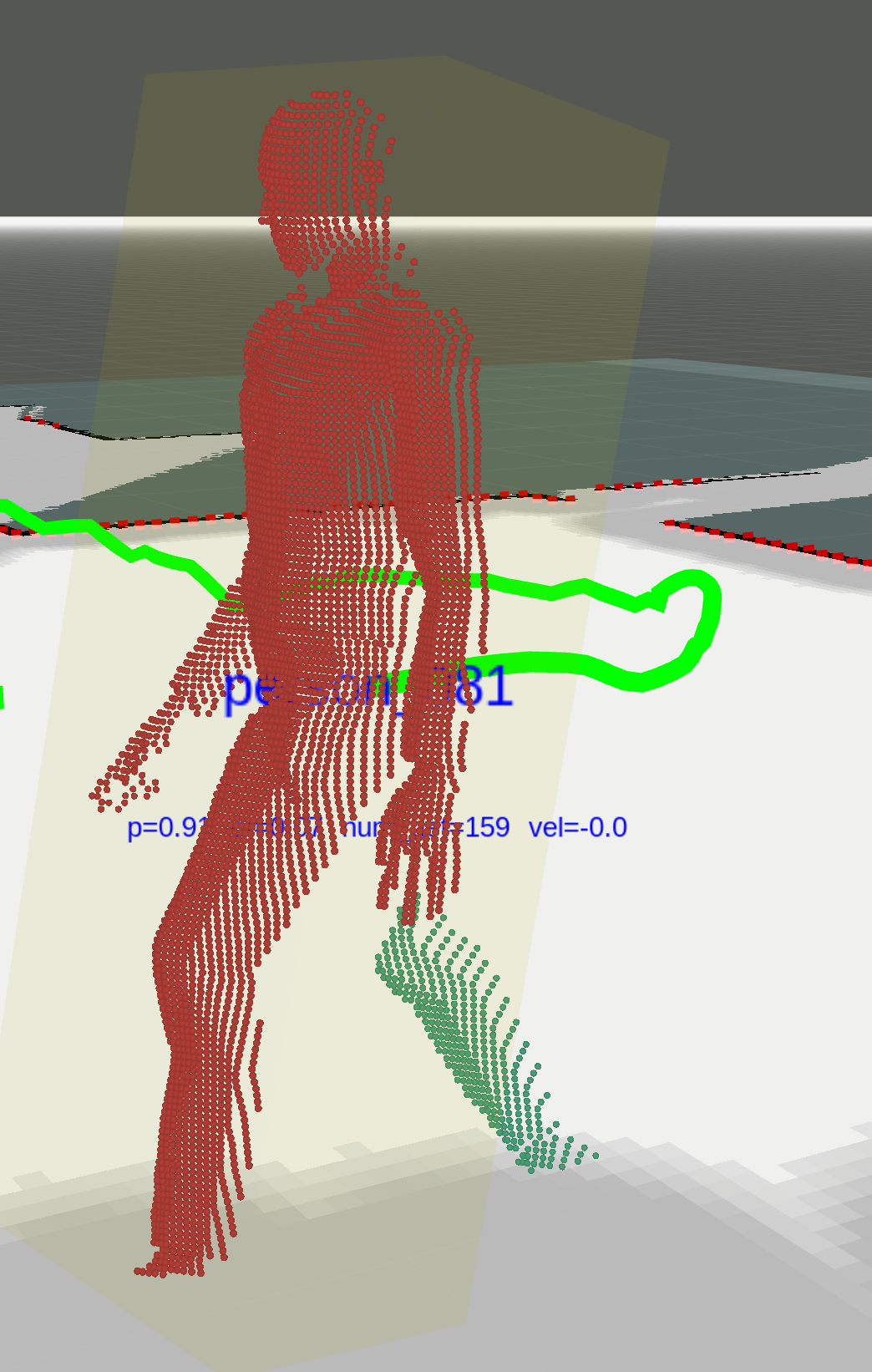}}; 

         \node[lnode,minimum width=1.1cm, minimum height=0.8cm] at (-3.1,-0.7)(mergeB) {Merge};     
         \path [draw, -latex] (segB) -- node[near end]{Structure}(mergeB);
         \node at (-5.9,-0.9)(segB_img){\includegraphics[width=0.09\linewidth, height=0.065\linewidth]{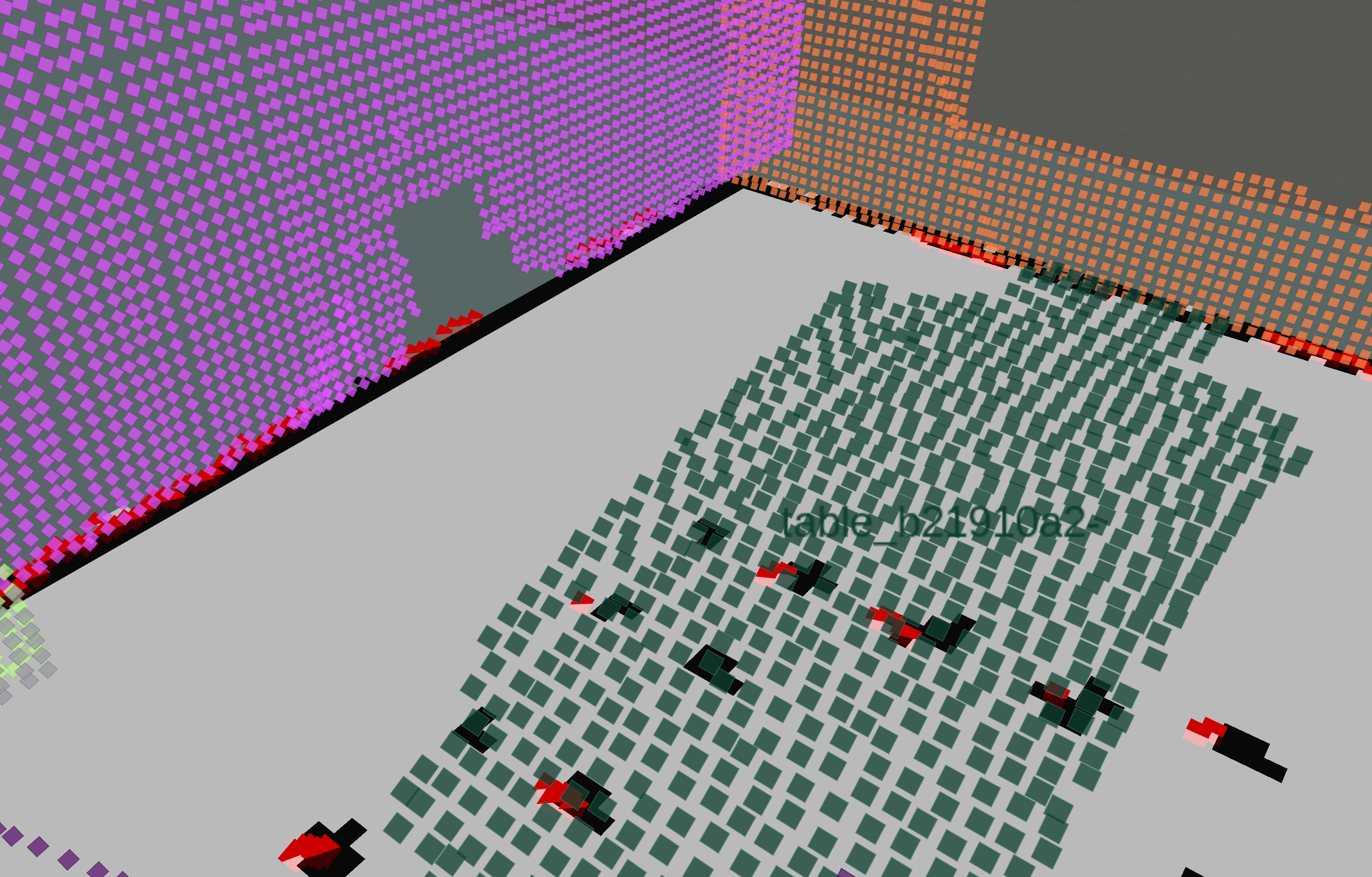}}; 

         \node at (-1.8,-0.9)(struc_img){\includegraphics[width=0.09\linewidth, height=0.065\linewidth]{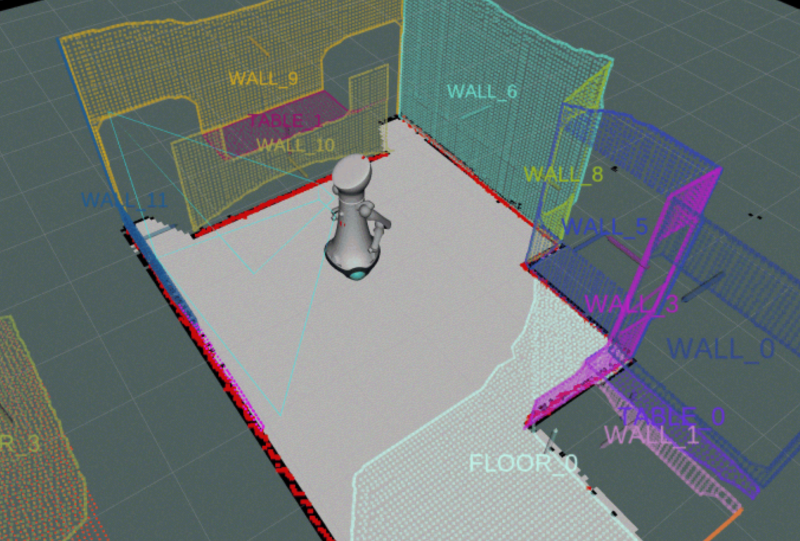}}; 
         \path [draw, -latex, dotted] (-3.9, -0.1) -- (mergeB);
         \path [draw, -latex,dotted] (-3.9, -0.1) -- (track);
         \node at (-4.4, -0.1){Pose};

         \path [draw,-latex] (matchF) -- node{$O_f$}(obs_img1);
     
         \node[align=center] at (1,0.1)(obj) {Foreground};
                  
         \node[rectangle, align=center] at (1,-0.7)(bstorage) {Background};
         \path [draw, -latex] (struc_img.east) |-  node[near end]{$O_b$, $I$}(bstorage);
         \node[rectangle] at (-0.4,0.9)(intersection) {};

         \path [draw] (obj_img) -- (intersection);
         \path [draw, -latex] (obs_img2) |- (obj)node[near start]{$O_f$, $I$};

         \node[rectangle, align=center] at (1,0.9)(env) {Env.};

         \node[lnode, minimum width=1.9cm] at (1,-1.5)(map) {Spatial Map};
         \path [draw, -latex] (slam) -|  node[near start]{Map 2D\&3D}(map);

         \node[lnode, minimum width=4.cm] at (1,-2.3)(dbs) {Service Interface};
        
         \node[lnode] at (4.2,0.5) (heat) {Heatmap \\ Generator};
         \node at (5.5,0.5)(heat_img){\includegraphics[width=0.08\linewidth, height=0.065\linewidth]{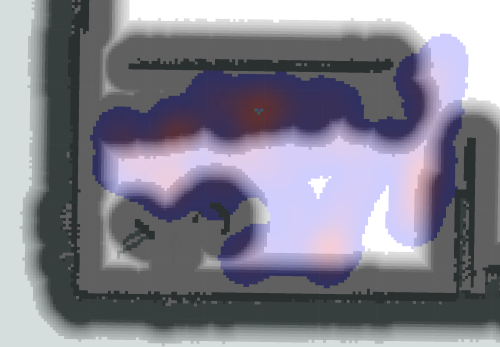}}; 

         \node[lnode] at (4.3,-0.5)(oa) {...};

         \node[lnode] at (4,2)(ml) {Feature \\ Learning};
         \node at (5.4,2.1)(ml_img){\includegraphics[width=0.09\linewidth, height=0.065\linewidth]{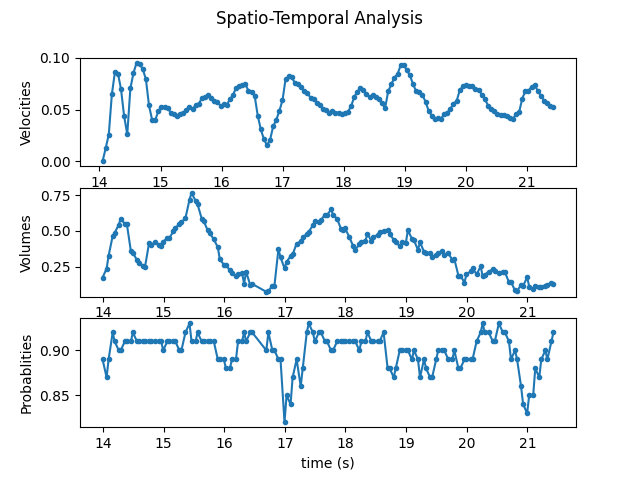}}; 

         \node[rectangle, align=center] at (2.1,0.9)(l1) {\textbf{Layer 1}};
         \node[rectangle, align=center] at (2.5,0.1)(l2) {\textbf{Layer 2}};
         \node[rectangle, align=center] at (2.9,-0.7)(l3) {\textbf{Layer 3}};
         \node[rectangle, align=center] at (3.3,-1.5)(l4) {\textbf{Layer 4}};

         \path [draw, latex-latex] (l2) --  node{$a$}(heat.west);

         \path [draw, latex-latex] (l2) --  node[near end]{$a$}(oa.west);
         \path [draw, latex-latex] (l2) --  node[near end]{$a$}(ml.west);

  
       \end{tikzpicture}
    }
    \caption{Schematic overview of the HIPer model with its newly developed~(blue) and reused~(gray) components. A triplet split separates recognition, consisting of a background and foreground pipeline to obtain observations~$o$ and instances~$i$, representation of this scene knowledge in multiple layers, and its interpretation for long-term scene analyses~$a$.}
    \label{fig:pipeline}
\end{figure*}
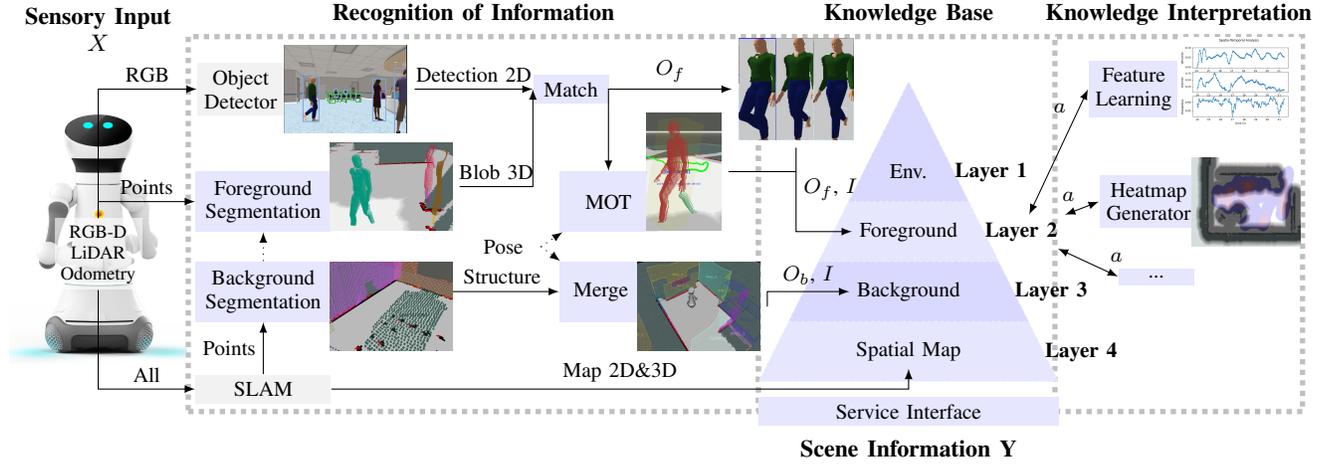

The proposed concept has been realized through a modular architecture, facilitating the seamless interchangeability of specific components. Subsequent sections provide details of each component, as illustrated in the schematic overview in Fig.~\ref{fig:pipeline}.

\section{Recognition of Information}
\label{sec:reco}
Using deep learning models became the standard in the open-world context as presented by the related approaches. However, their usage is challenging for the addressed perception of arbitrary objects since they cannot be easily split, merged, and reused for other than the trained purpose. When comparing DL-based object detectors and semantic segmentation, the last would be more challenging when facing a new task due to their lower availability and smaller number of object classes.

The \gls{acro:HI} provides a hybrid approach of multiple modules, combining DL-based and traditional methods. The recognition leverages its performance by reusing a 3D-SLAM and an object detector, two comprehensive research areas with a wide selection of methods for many applications and domains. 
The background reconstruction uses the SLAM's output to further process the registered points with a semantic segmentation based on traditional techniques to obtain primitive structures. This procedure allows for precise reconstruction of static regions of the scene while requiring little computation power. The foreground recognition comprises a segmentation component that uses the background as a preattentive filtering, the detector that delivers object proposals, a matching to obtain 3D observations~$O_f$ based on the IOU, and finally, a MOT for its aggregation into instances~$i_b$.

\subsection{Sensory Input}
Humans employ various senses to obtain environmental information, allowing for a more comprehensive scene perception since deficiencies in the different principles can be compensated. Due to the richness of information, humans rely 80-\SI{85}{\percent} on visual recognition~\cite{ripley2010}. However, balancing information and computation power is essential for mobile systems. Therefore, the \gls{acro:HI} mimics this approach by concentrating on visual perception using an RGB-D camera. Supplementary sensors, a 2D LiDAR and wheel odometry, are integrated to improve the robot's localization capabilities.

\subsection{SLAM}
\gls{acro:SLAM} provides a diverse range of established methods. When reusing an existing approach, a thorough performance comparison is necessary. The performance of prevalent feature-based \gls{acro:3D}-\gls{acro:SLAM} methods is discussed in~\cite{Jesus2021}. Approaches like \emph{ORB-SLAM2}, \emph{RTAB-Map}, and \emph{SPTAM} demonstrate an equally outstanding performance with slight variations depending on factors like the environment~(indoors or outdoors) and choice of sensor(s). Exemplarily, RTAB-Map~\cite{Labbe2018}, a widely-used graph-optimized dense \gls{acro:SLAM} method, is chosen due to its popularity. RTAB-Map employs advanced techniques to robustly reconstruct the spatial scene in \gls{acro:3D} and is interchangeable since alternative methods share equivalent inputs and outputs. The output of the SLAM, illustrated by the reused component in Figure~\ref{fig:pipeline}, includes registered point clouds with the graph's pose ID, enabling anchor observations~$o$ on this graph while maintaining awareness of loop closures and post-optimizations to update aggregated instances.

\subsection{Background Segmentation}
Preliminary tests indicate that most background data within indoor environments pertains to building structures, which can be defined by basic shapes. Since the background is assumed to be semi-static, the segmentation uses the registered point cloud on the SLAM's graph as input to obtain~$O_b$, as summarized in Algorithm~\ref{alg:extract_planes}. First, the point cloud is downscaled to extract planes efficiently using \emph{SACSegmentation}~\cite{Rusu2011}. Afterward, a simple plane classification based on the alignment of the normal vector assigns the object class: \emph{wall}, \emph{floor}, \emph{ceiling}, and \emph{table} or \emph{unknown}.
\begin{algorithm}[tb]
\caption{Proceeding of extracting new background observations in 3D point clouds.}\label{alg:extract_planes}
\begin{algorithmic}[1]\small
\Function{ExtractOb}{$points$, $pose$}
\State $points$ = \textproc{DownscalePoints}($points$) \Comment{reduce the resolution}
\State $points$ = \textproc{TransformToRobot}($points$) 
\State $planeList$ = \textproc{DetectPlanes($points$)} 
\State $planeListClassified =  $\textproc{Classifier($planeList$)}
\For{$plane$ \textbf{in enumerate} $ planeListClassified$} 
    \If {$plane.Type$!=UNKNOWN}
        \State $O_b[plane])$ = \textproc{GenerateAndUpload}($plane$)
    \EndIf
\EndFor
\State \textbf{return} $O_b$
\EndFunction
\end{algorithmic}
\end{algorithm}
Finally, $o_b$ is generated from the plane and uploaded to the knowledge base. In this step, meta-information, such as an \gls{acro:UUID}, an aligned bounding box, the current robot pose id, and the contour of the plane, are extended.

\subsection{Background Merging}
The acquired observations~$O_b$ correspond to either a previously or newly perceived instance. Algorithm~\ref{alg:merge_o} describes the merging of corresponding observations with instances. 
\begin{algorithm}[tb]
\caption{Merging of background observation $o_b$ with instances $I_b$.}\label{alg:merge_o}
\begin{algorithmic}[1]\small
\Require $I_b$ \Comment{background instances from KB}
\Function{MergeOb}{$o_b$, $I_b$}
     \State $o_b$ = \textproc{TransformToMap}($o_b$, $\textproc{pop}(\textit{posegraph})$) 

    \State def $mergeList$
    \For{$i_b$ \textbf{in} $I_b$}    
        \State $mergeList$.add($i_b$) = \textproc{MergeCriteriaCheck}($o_b$, $i_b$)
    \EndFor
   \Switch {size($mergeList$)}
    \State \textbf{case} 0:
    \State \quad $o_b \rightarrow \text{NEW} \ i_{\text{new}}$
    \State \textbf{case} 1:
    \State \quad \text{MERGE} $o_b$ \text{to} $i_{\text{index}}$
    \State \textbf{case} \textgreater 1:
    \State \quad \text{MERGE} $o_b$ \text{to} $i_{\text{index}}$, \text{delete duplicates}
    \EndSwitch
\EndFunction
\end{algorithmic}
\end{algorithm}
First, each $o_b$ undergoes a transformation from the robot graph to the global 3D space. Subsequently, a set of criteria with increasing complexity is employed to assess the similarity of $o_b$ with each $i_j$, including class, alignment, bounding box, and finally, the contour by overlap. Upon success, $i_j$ is updated, i.e., the UUID of~$o_b$ is linked, and the spatial data are merged. Moreover, any potential duplicates are removed. In case of unsuccessful merging, $o_b$ is converted into a new instance.

\subsection{Object Detector}
Selecting a suitable object detector requires comparing their performance. The COCO dataset~\cite{Lin2014} offers common object classes found in the target application and is widely used for comparing performance. Thus, we selected three detectors with the highest mAP\footnote{\url{https://paperswithcode.com/sota/object-detection-on-coco}, accessed 01.12.2023}: \emph{YOLO V6}~\cite{Li2022}, \emph{DETR}~\cite{Carion2020}, and \emph{SWIN~V2}~\cite{Liu2022}. Table~\ref{tab:object-detectors} compares their performance with runtime-oriented settings. Although the method \emph{SWIN V2} achieves the highest precision, followed by \emph{DETR}, \emph{YOLO V6} performs significantly faster and more efficiently, as indicated by the frame rate and inference speed. Considering this limitation, we propose using \emph{YOLO V6} since it achieves the best performance with respect to the availability of large and diverse datasets, exchangeability, and computation effort.
\begin{table}[tb]
\caption[Object detector performance comparison]{Performance comparison of object detector with the highest mAP for the COCO dataset~(*3rd party results). The frame rate and inference speed are measured with the same processing unit. The video input is set to 640x480 pixels at \SI{30}{Hz}.}
\centering
\small\renewcommand*{\arraystretch}{1.2}
\begin{tabularx}{0.99\linewidth}{p{\dimexpr.34\linewidth-2\tabcolsep-1.3333\arrayrulewidth}XXX}

\toprule
\textbf{Description} & \textbf{YOLO V6} & \textbf{DETR} & \textbf{SWIN V2}  \\
\midrule
Frame Rate (Hz) & 29.8 & 6.79 & 10.5 \\
Inference Speed (CPU; GPU) & 240; 21 & 275; 44 &  340; 10\\
Detection Accuracy (mAP*) & 58.2 & 64.2 & 65.0 \\
\bottomrule
\end{tabularx}
\label{tab:object-detectors}
\end{table}

\subsection{Foreground Segmentation}
The foreground segmentation aims to extract region proposals of instances by obtaining a set of observations $O_f$ each time the RGB-D camera provides new data as summarized by Algorithm~\ref{alg:extract_foreground}. The approach enhances its performance by removing the background to narrow down the regions of interest. This strategy is akin to human preattentive filtering~\cite{Graf2022a}, enabling the processing to focus attention on areas that exhibit greater complexity in terms of visual properties and dynamics compared to the background. Foreground objects are assumed to be dynamic, necessitating the consideration of object motion and the robot's ego-motion simultaneously, supported and referenced by the background. Specifically, bounding boxes of initially mapped scene structures are filtered from the high-frequency point cloud of the robot camera. The remaining points are then 3D segmented into regions without semantics using Euclidean Cluster Extraction~\cite{Rusu2011}. Finally, the observations are generated equivalent to those of the background.
\begin{algorithm}[tb]
\caption{Implemented foreground object proposal segmentation.}\label{alg:extract_foreground}
\begin{algorithmic}[1]\small
\Require $points$ \Comment{point cloud from camera}
\Require $I_{b}$ \Comment{sum of background instances from KB}
\Function{SegmentForeground}{\textbf{input} $points$}
\State $points$ = \textproc{DownscalePoints}($points$) \Comment{reduce the resolution}
\State $I_b.bbox \leftarrow $  \textproc{loadBackground} \Comment{get bounding boxes}
\State $I_{b}.bbox$ = \textproc{TransformToRobot}($I_{b}.bbox$)
\State $points$ = \textproc{FilterBackground}~($points$, $I_{b}.bbox$) \Comment{remove points from background}
\State $regionList$ = \textproc{SegmentPoints}~($points$) \Comment{using EuclideanClusterExtraction}
\For{$region$ \textbf{in enumerate} $regionList$}
    \State $O_f[region])$ = \textproc{GenerateAndUpload}($region$)
\EndFor
\State \textbf{return} $O_f$  
\EndFunction
\end{algorithmic}
\end{algorithm}

\subsection{Foreground Matching}
In parallel to the foreground segmentation, the previously selected detector uses the color image of the camera to detect objects. The matching tries to assign the semantic to the region proposals as presented in the Algorithm~\ref{alg:assignment_foreground}. First,  it synchronizes detections and 3D blobs using their time stamps. Then, the \gls{acro:IOU} is calculated after projecting the 3D blobs into the image plane. 
If the \gls{acro:IOU} crosses a certain threshold, the object class is extended to $o_f$.
Although this approach allows handling many objects, unknown objects must be considered in real-world applications. Thus, not matched regions are tagged as unknown. 
\begin{algorithm}[tb]
\caption{Semantic assignment of 3D-blobs proposals with bounding boxes.}\label{alg:assignment_foreground}
\begin{algorithmic}[1]\small
\Require $O_f$ \Comment{3D blob from foreground segmentation}
\Require $detectionList$ \Comment{bounding boxes from detector}
\Ensure~($O_f$, $detectionList$)  $\leftarrow$ synchronized
\Function{AssignSemantic}{\textbf{input} $O_f$, $detectionList$}
    \State $IOU  \leftarrow \text{EmptyMat}(\text{length}(O_f), \text{length}(detectionList))$
    \For{$o_{f}$ \textbf{in enumerate} $O_f$}
        \For{$d$ \textbf{in enumerate} $detectionList.\text{box}$}
            \State $IOU_{image} = \textproc{CalculateIOU}(o_{f}.\text{region}, d.\text{box})$ 
            \If {$IOU_{image}OU > threshold$}
                \State $o_{f} \leftarrow d.\text{objectclass}$ \Comment{semantic assignment}
            \EndIf            
        \EndFor
    \EndFor
\EndFunction
\end{algorithmic}
\end{algorithm}

\subsection{Multi-Object Tracking}

The foreground observations $o_f$ are fed into a real-time optimized \gls{acro:MOT} to merge multiple 3D detections into instances~$i$ with the consideration of dynamics. The approach uses a Hungarian algorithm\cite{Kuhn1955} for matching the object center in world coordinates. Afterward, a Kalman Filter~\cite{Thrun2005} tracks each object, using its position and derived velocity. The predicted velocities without updates are continuously reduced until zero to avoid shifting outside map boundaries.

\section{Knowledge Representation}
\label{sec:rep}
The knowledge base represents acquired scene information in a known structure. In this context, human knowledge consists of semantic and episodic memories obtained through life-long learning, organized hierarchically and stored in multiple spatial layers~\cite{Hommel2000, Wagemans2014}. The \gls{acro:HI} adapts this concept but already benefits from initial knowledge deployment and simplified data exchange between multiple agents~\cite{Graf2022a}. Prior knowledge is deployed through training of the detector for knowing relevant semantics in the particular environment.

\subsection{Knowledge Structure}

Adapting the hierarchical structure from human perception implies layered modeling of available scene data, anchoring with the consideration of updates after graph optimizations, and handling episodic memories that link to the corresponding observations. Related work uses two fundamental concepts of data management systems: \gls{acro:RDBMS} and document-oriented databases~\cite{Graf2022a}. RDBMS require explicit modeling of knowledge through a predefined ontology, while document-oriented databases merely require implicit modeling without any restrictions, allowing the linking of documents to build an ontology.  
For the addressed cross-task application, we emphasize using a document-oriented database over RDBMS as it offers the required flexibility, e.g., having different expert knowledge but the same common understanding), for domain transfer, and in-operation extensions. Vice versa, RDBMS are beneficial for complex knowledge modeling such as the \emph{Knowrob}~\cite{Beetz2018}.
As ontology, related approaches, such as \emph{Kimera}~\cite{Rosinol2021} and \emph{Hydra}~\cite{Hughes2022}, respectively, model hierarchical semantic structures of buildings linked through scene graphs. The \gls{acro:HI} sets up  the document-oriented database with a similar spatial structure; however, it splits scene knowledge into four layers according to Fig.~\ref{fig:pipeline}: environment ID, foreground, background, and spatial map. This flat hierarchy was chosen to maximize flexibility for cross-applications and cross-robot use while supporting their ontology for certain scenarios. Our database stores each scene information in a separate document. Documents of instances link to corresponding observations through their UUID to define an implicit relation between them.

\subsection{Entities}
The knowledge base stores $O$, $I$, and $A$ according to Equation~\eqref{eq:E}. Each entity has a hierarchical structure comprising redundant high-level and individual low-level attributes. High-level attributes are the UUID, the corresponding robot frame name, the time of the last update, the object class, and the bounding box. While instances are linked to the global map frame, observations are linked to the sensor frame, allowing recalculating instances with updated observations pose after graph optimizations of the SLAM. Low-level attributes of observations are raw sensor data (e.g., masked color image) within the Region of Interest~(ROI). In contrast, low-level attributes of instances are aggregated spatial data (e.g., fused points) with a trajectory and links to a batch of time-continuous child observations. In addition, general data, such as the robot trajectory, spatial map data, and cross-class properties, are stored. 

\subsection{Accessibility of Scene Knowledge}
The service interface provides scene information for decision-making, navigation, manipulation, and scene visualization. In particular, SQL queries are used to search instances~$i$ in the database based on their attributes (e.g., UUID, object class), or a criterion, such as \texttt{closest table}, or \texttt{cups in the FOV}. 
Motion planning for manipulation requires calculating the distance between each part of the robot arm and each obstacle simultaneously, thus a small amount of data points is recommended. This is facilitated by an additional interface that retrieves scene data within a specific workspace that can be activated and deactivated. In particular, it provides a static background, represented through bounding boxes with semantics. In parallel, it provides the dynamic foreground, represented through an occupancy grid with semantics for object-aware obstacle avoidance to the manipulation. 

\section{Knowledge Interpretation}
\label{sec:inter}

The interpretation of scene knowledge~$A$ encompasses multiple dimensions, including time, space, and relations, allowing for a comprehensive scene analysis over an extended period. These techniques enhance scene understanding by extending and robustifying object instances~$i$ and scene semantics during operation within the scene. Exemplarily, two features exhibiting high potential for multifunctional robots are incorporated: feature learning and heatmaps.

\subsection{Feature learning}\label{learning}
Learning environment-specific perceptual information is a major step toward human-like life-long learning. The objective is to determine environment-specific object properties, acquired through a time-series analysis of a particular instance or all instances of a certain class. The typical object dynamic, volume, and altitude are determined since they represent valuable information to understand their behavior.
\subsubsection{Object Dynamic} To determine the object dynamic of a particular class, their trajectories are split into two sets: static~$S$ and dynamic~$D$. Transitioning between these states is less clear due to expected measurement noise during the stationary phase. An \gls{acro:NDT} of the measurement noise is expected. The absolute velocity is considered, thus, a folded \gls{acro:NDT} with zero mean~$\mu$ and a variance~$\sigma^2$ describes the probability density function of the stationary portion as
\begin{equation}
S_{prob}(|\vec{v}|; \sigma) = \sqrt{\frac{2}{\pi}} \frac{1}{\sigma} e^{-\frac{|\vec{v}|^2}{2\sigma^2}}.
\end{equation}
The threshold selected for the stationary to dynamic transition is $2 \sigma$, i.e., at \SI{95.45}{\percent} of the noise. The dynamic portion below this threshold is described with the inverse noise distribution. Above, due to no prior object knowledge, a uniform distribution was selected to describe the dynamic portion. However, a class-specific dynamic range is expected within the confidence interval $CI_D = [c_l; c_u]$. The resulting probability density function of the dynamic portion is described as
\begin{equation}
D_{prob} (|\vec{v}|; \sigma_{\varepsilon}^2) =
\begin{cases}
1-e^{-\frac{\vec{v}^2}{2\sigma_{\varepsilon}^2}}, & \text{if } |\vec{v}| < 2 \sigma \\
1, & \text{if } 2 \sigma < |\vec{v}| < \text{97.5th perc.}.
\end{cases}
\end{equation}
The \emph{Object Dynamic Score}~($ODS$) is introduced, calculated for each object class as 
\begin{equation}
ODS= [\mu_{\vec{v}}, c_l, c_u, r].
\end{equation}
For each class, it indicates the mean velocity $\mu_{\vec{v}}$, the confidence interval of the velocity, and a dynamic-to-static ratio~$r$, which allows reasoning about the object class reactivity. We used the object dynamic to adapt the MOT pipeline in two ways: First, optimize the tracking by using $\sigma$ to calculate class-specific covariances, and second, reject distances of the Hungarian algorithm beyond~$c_u$ multiplied by the time delta~(condition: dt~$<$~\SI{1}{\second}).

\subsubsection{Object Size} Determining this feature is essential to understanding the typical size of objects in the scene, particularly important since the detector uses \gls{acro:2D} data without depth measurements. To minimize false positives, sizes beyond an experienced range are filtered. The distribution of the bounding box volumes is considered under the assumption of no shape change since it is invariant to origin and orientation. Similar to the object dynamic, a uniform distribution is selected. Bootstrapping calculates the \SI{95}{\percent} confidence interval~($CI$) for class-specific bounds, predicting future sizes fit within this range with the same probability.

\subsubsection{Object Altitude} Determining an object's height above the floor follows the same methodology as the object size. The altitude is calculated by subtracting the vertical bounding box size from the object's center. The confidence interval of the resulting uniform distribution helps minimize false positive detections, such as understanding that people walk on the floor altitude, enabling optimization of robot behavior based on anticipated object locations.

\subsection{Heatmap generator} Learning the locations of a particular instance or object class is valuable to understanding regularities since it provides feedback about frequently observed areas of a specific time. The knowledge base provides historical scene data to generate so-called heatmaps with the procedure in Algorithm~\ref{alg:heatmap}, \added{an improved version first introduced in}~\cite{Graf2022b}. Instance data within a specified time delta is obtained from the knowledge base using a query with the desired start and end time and the instance type. It inflates instance trajectories with a normal distribution. The heatmaps are filled up to an upper threshold \SI{80}{\percent} occupancy and overlayed with the navigation costmap to avoid particular areas, such as crowded spaces. Alternatively, their inverse representation can be generated to explore these regions. 
\begin{algorithm}[tb]
\caption{Proceeding of the heatmap generation.}\label{alg:heatmap}
\begin{algorithmic}[1]\small
\Require $I_f$ \Comment{sum of relevant instances from KB}
\Require $t_1$, $t_2$ \Comment{start and end time}
\Require $map$ \Comment{2D grid map}
\Function{HeatmapGenerator}{\textbf{input} $map$}
\State $pointList$, $\mu_{\vec{v}}$ = \textproc{QueryKB}($I_f$, $t_{start}$, $t_{end}$) 
\State $heatmap$ =  $map$.copy
\For{$point$ \textbf{in} $pointList$}
    \State $heatmap.at[point]  \leftarrow$ NDT~($\mu_{\vec{v}}$)  \Comment{overlay NDT}
\EndFor
\State $heatmap  \leftarrow$ scaled
\EndFunction
\end{algorithmic}
\end{algorithm}
\section{Experiments}
\label{sec:eval}
The experiments aim to evaluate the performance of holistic scene understanding and its benefits for multifunctional mobile robots within a long-term running setup. 
Unfortunately, we could not find such a benchmark, where a mobile robot repeatedly perceives a restricted area with dynamic objects. Related work uses specific datasets of one-time routes, e.g. in~\cite{Rosinol2021}, or non-reproducible real-world settings~\cite{Kunze2018, Suchan2017, Galindo2005}, aiming to quantify a specific part of perception, such as the metric-semantic reconstruction accuracy. Given the objective of a holistic approach and the fact that it integrates well-known recognition techniques, i.e., SLAM and an object detector, existing benchmarks are unsuitable. Therefore, a new benchmark has been developed to measure the perception's performance and its implications for typical tasks in a long-term running setup. The benchmark uses a single-setting ablation study to reveal the impact of each component on the overall performance. Noteworthy, simulated, and real-world environments are supported to have a reproducible baseline and to verify a sim-to-real transfer, respectively. 

\subsection{Scenario and Setup}
The scenario and setup of the \gls{acro:HI} address multi-functional robots in the open-world context that repeatedly take over various arbitrary tasks. Therefore, the scenario choice and design must be according to this. 

\subsubsection{Scenario} 
The scenario, covering the robot's tasks, includes essential techniques and physical interactions with the environment. While collecting and carrying objects can be tedious for humans, it is a manageable task for robots and a popular function of service robots~\cite{Bajones2015, Odabasi2022}. User studies, such as in~\cite{Yamamoto2019}, highlight this task as crucial in supporting humans in domestic environments. Therefore, a \emph{fetch-and-carry} scenario is selected consisting of four sub-tasks, abbreviated as~T, that address significant capabilities:

\begin{description}[itemsep=5pt]
\item [T1] \emph{Find Person}: Start at an idle position and search for a person to offer help. The task ends when the person has been detected.
\item [T2] \emph{Search Object}: After detection, the person is assumed to request a bottle which is subsequently searched for by the robot. The task ends when a bottle has been detected.
\item [T3] \emph{Grasp Object}: Start at the position where the bottle has been detected. Approach and grasp the bottle.
\item [T4] \emph{Carry Object}: Carry the bottle to the idle position.
\end{description}
The first task tackles the robot's navigation and ability to find a dynamic object, i.e., a person in the scene. The second task is similar, but the target object is semi-static. The third task uses the robot's manipulation to grasp the object. The last task is a pure transport task from point A to B, where the robot has to efficiently and safely navigate to its goal.

\subsubsection{Setup}
To perform these tasks, we selected three different everyday environments. While two simulated environments provide a reproducible and known baseline, one real-world environment demonstrates a realistic setup without a fully known baseline. Both share the same notebook as the processing unit\footnote{CPU: i9-11980HK, GPU: Nvidia 3060 RTX, RAM: 32GB, OS: Ubuntu 20.04 with ROS noetic} and use an RGB-D camera with 1280x720 pixels at \SI{30}{Hz} to allow a comparison of the perceptual performance.

\begin{figure}[tb]
  \centering

\begin{subfigure}{.5\linewidth}
  \centering
  \begin{tikzpicture}

    \node at (0.0, 0.0) {  \includegraphics[height=1.1\linewidth]{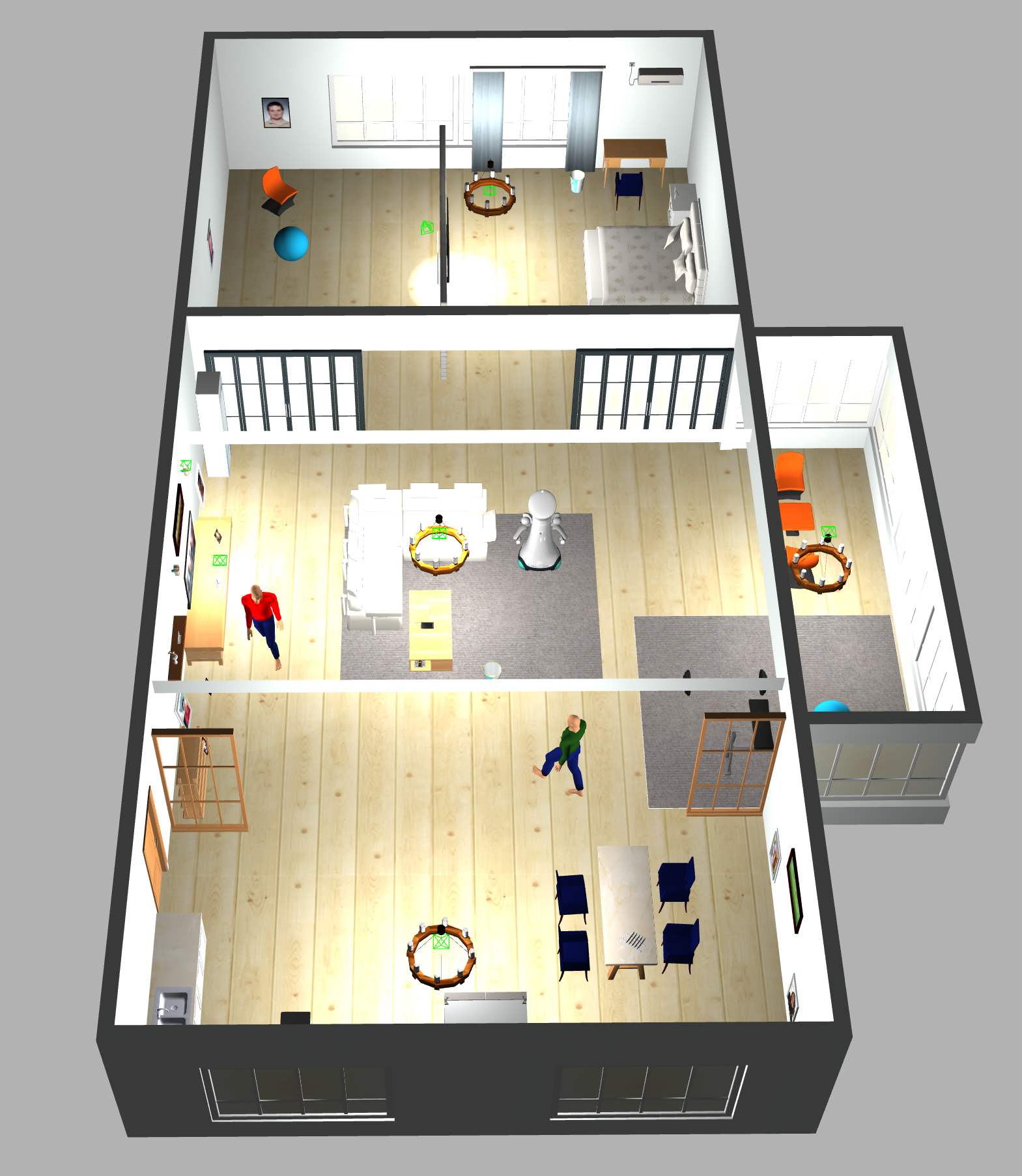}};

    \node[fill=white!30, rounded corners, inner sep=5pt](idle) at (1,2) {Idle position};
    \node[fill=white!30, rounded corners, inner sep=5pt](bottle) at (0.4,-0.5) {Bottle};
    \fill[red] (-0.4, 1.6) circle (0.1); 
    \fill[red] (0.4,-1) circle (0.1); 
    \draw[->] (idle.west) --  (-0.4, 1.6);
    \draw[->] (bottle.south) -- (0.4,-1);

    \end{tikzpicture}
    \caption{House}
    \label{fig:house}
\end{subfigure}%
\begin{subfigure}{.5\linewidth}
  \centering
    \begin{tikzpicture}

    \node at (0.0, 0.0) {\includegraphics[height=1.1\linewidth]{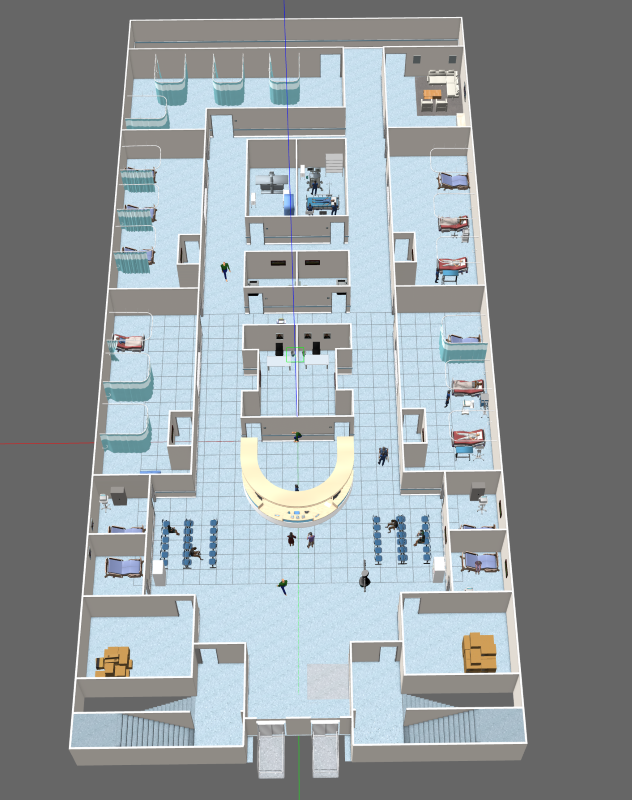}};

    \node[fill=white!30, rounded corners, inner sep=5pt](idle) at (0.8,2) {Idle position};
    \node[fill=white!30, rounded corners, inner sep=5pt](bottle) at (-0.4,-1.6) {Bottle};
    \fill[red] (-0.8, 1.5) circle (0.1); 
    \fill[red] (-0.4,-0.5) circle (0.1); 
    \draw[->] (idle.west) -- (-0.8, 1.5);
    \draw[->] (bottle.north) -- (-0.4,-.5);
    \end{tikzpicture}
    
  \caption{Hospital}
  \label{fig:hospital}
\end{subfigure}
\caption{For evaluation selected publicly available virtual environments from Amazon~\cite{AWS_Gazebo}. Walking people are spawned to enrich dynamics.}
\label{fig:envs}
\end{figure}
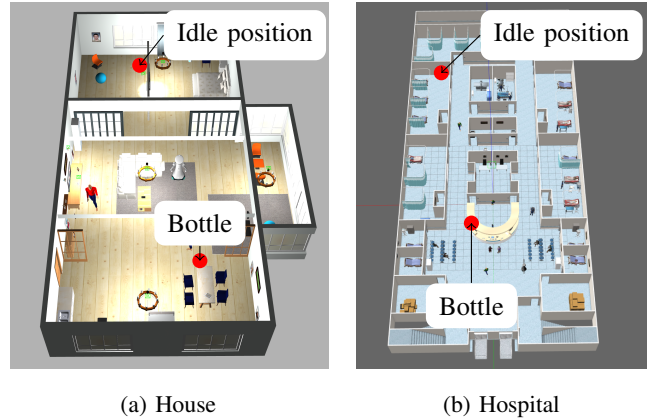
\begin{figure}[tb]
  \centering
\begin{subfigure}{.35\linewidth}
  \centering
  \includegraphics[width=0.95\linewidth]{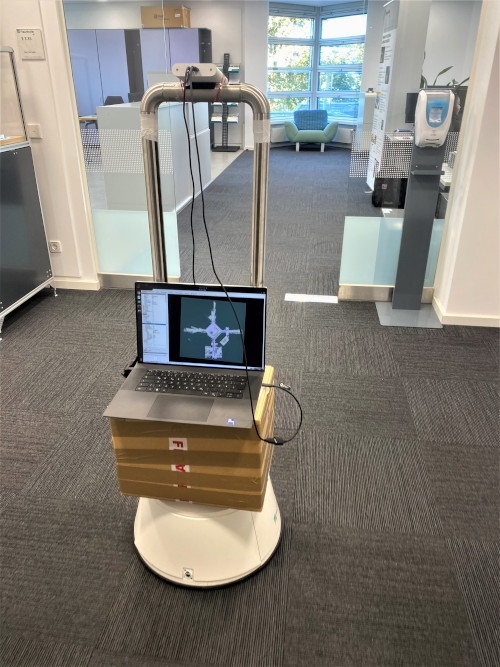}
  \caption{MobiKa}
  \label{fig:robot}
\end{subfigure}%
\hfill
\begin{subfigure}{.625\linewidth}
  \centering
  \includegraphics[width=0.95\linewidth]{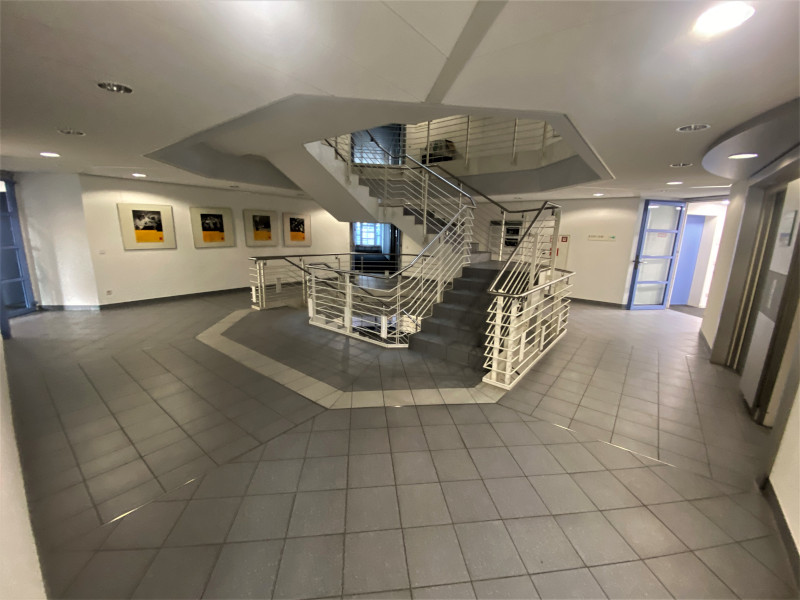}
  \caption{Office Building}
  \label{fig:office}
\end{subfigure}
\caption{Real-world office setup of the robot \emph{MobiKa} (a), comprising a Kinect Azure RGB-D camera and notebook mounting. \added{The experiments take place in an office building (b) comprising a central stairwell and a lab connected by corridors.}}
  \label{figures:realenv}
\end{figure}

The \gls{acro:HI} and related work share the use of the Robot Operating System~(ROS). Thus, we emphasize taking its \emph{Gazebo} simulator over \emph{Unity}, used by~\cite{Rosinol2021}, \emph{Unreal Engine}, or \emph{AI Habit},  due to its native support of the entire robot data. The simulated experiments use two popular open-source Gazebo worlds, house, and hospital~(see Fig.~\ref{fig:envs}), provided by Amazon~\cite{AWS_Gazebo}. The virtual experiments are performed with \emph{Care-O-bot~4}, a multifunctional mobile service robot~\cite{kittmann2015}. Both virtual environments are static, whereas dynamic objects have been integrated to achieve a more realistic setup, particularly important for evaluating foreground recognition and spatio-temporal scene analysis. Since people represent the most important dynamic objects for service robots, animated people have been added using the Social Force Model~(SFM) proposed in~\cite{Helbing1995}. One person is moving randomly in the house and five people are moving between predefined positions in one corridor of the hospital. 

The real-world experiments take place in an office building~(Fig.~\ref{fig:office}) using our research robot \emph{MobiKa}~\cite{Graf2019}~(Fig.~\ref{fig:robot}). The robot is fully controlled with the same notebook as used for simulation. We neither adapted the environment nor instructed people how and where to move to achieve a realistic setup.

\begin{table}[tb]
    \centering
    \renewcommand*{\arraystretch}{1.3}
    \small
    \newcolumntype{C}{>{\centering\arraybackslash}X}
    \caption{Selected ablation study configurations to identify the component-wise contribution to the overall performance.}
    \begin{tabularx}{0.95\columnwidth}{CCCCCCC}
        &
        \rotatebox[origin=l]{45}{\textbf{Basic Model}} & \rotatebox[origin=l]{45}{\parbox{\linewidth}{\raggedright\textbf{Foreground}\\\textbf{Recognition}}} & \rotatebox[origin=l]{45}{\parbox{\linewidth}{\raggedright\textbf{Background}\\\textbf{Recognition}}} & \rotatebox[origin=l]{45}{\textbf{Knowledge Base}} &  
        \rotatebox[origin=l]{45}{\textbf{Heatmaps}} & 
        \rotatebox[origin=l]{45}{\parbox{\linewidth}{\raggedright\textbf{Feature}\\\textbf{Learning}}} \\
       \hline
        \textbf{C1} & \multicolumn{1}{C}{x} & x & x & x   & x & x \\
        \textbf{C2} & \multicolumn{1}{C}{x} & x & x & x   & x &   \\
        \textbf{C3} & \multicolumn{1}{C}{x} & x & x & (x) &   &   \\
        \textbf{C4} & \multicolumn{1}{C}{x} & x &   & (x) &   &   \\
        \textbf{C5} & \multicolumn{1}{C}{x} &   &   &     &   &   \\
      \bottomrule
    \end{tabularx}
    \label{tab:ablationconfig}
\end{table}

\subsection{Experiment Execution}
The tests are split into the initial mapping and its following long-term operation for each environment. The initial mapping only uses the background recognition pipeline at \SI{2}{Hz} without scene dynamics, i.e., without walking people, a common approach in robotics to reconstruct the scene by a systematic, user-controlled exploration of the entire environment. Afterward, the SLAM switches to localization mode. Walking people were spawned as described in the setup when starting long-term operations. The robot systematically covers the scene for two hours with an implementation presented in~\cite{Bormann2018}. The single-setting ablation study then assesses how each component contributes to the scenario performance. T1-T4 are performed with a sample size of ten per environment and configuration as defined in Tab.\ref{tab:ablationconfig}.  One exception concerns the skipping of T3 in the real-world experiment since the robot has no manipulator.


\subsection{Results}
The results are split into background reconstruction and long-term perception to obtain the perceptual performance, followed by the ablation study.

\subsubsection{Background Reconstruction}
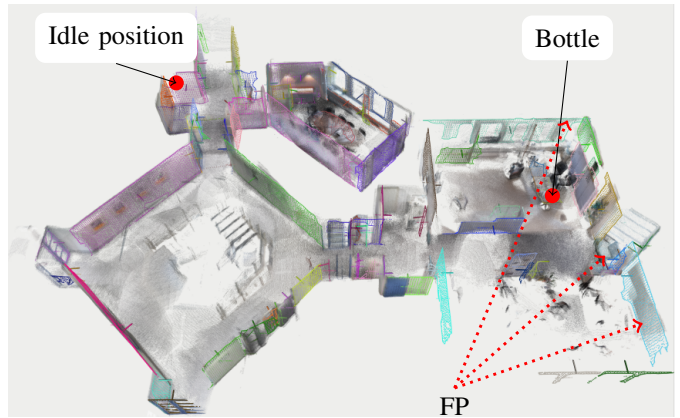
\begin{figure}[htb]
    \begin{tikzpicture}
    \node at (0.0, 0.0) {\includegraphics[width=1\linewidth]{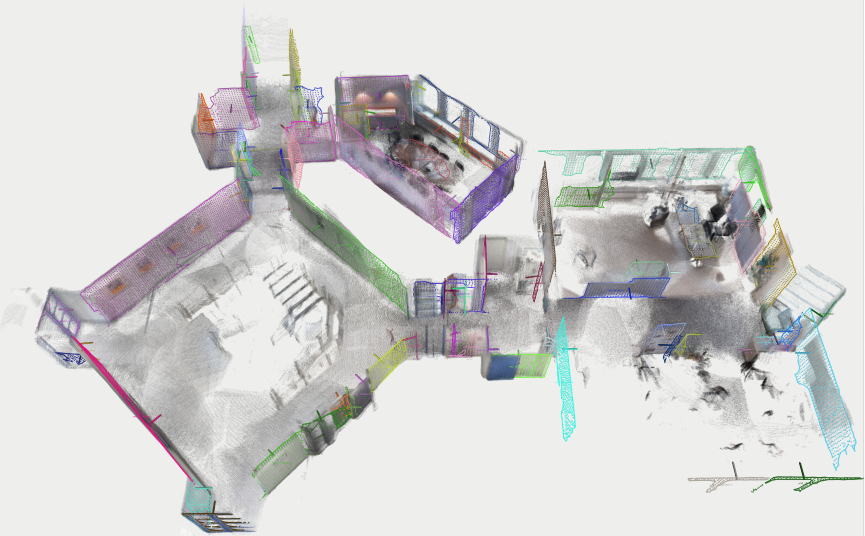}};

    \node[fill=white!30, rounded corners, inner sep=5pt](idle) at (-3,2.3) {Idle position};
    \node[fill=white!30, rounded corners, inner sep=5pt](bottle) at (3,2.3) {Bottle};
    \fill[red] (-2.2, 1.7) circle (0.1); 
    \fill[red] (2.8,0.2) circle (0.1); 
    \draw[->] (idle.south) --  (-2.2, 1.7);
    \draw[->] (bottle.south) -- (2.8,0.2);
     \node (fp) at (1.5,-2.6) {FP};
    \draw[->, dotted, red, line width=1.1pt] (fp.north) -- (3.5, -0.6);
    \draw[->, dotted, red, line width=1.1pt] (fp.north) -- (4,-1.5);
    \draw[->, dotted, red,  line width=1.1pt] (fp.north) -- (3,1.2);
    \end{tikzpicture}
    
\caption{Reconstructed office background. Whereas static parts have been assigned successfully throughout, the semantic classification fails for large furniture, such as for shelves, since they have been labeled as walls~(FP). The idle and bottle positions represent the key positions for the scenario.}
\label{fig:reconstruction}
\end{figure}

The background reconstruction results in Tab.~\ref{tab:backgroundsemantics} indicate a precision \SI{>98}{\percent} in the static-dynamic split with only a dynamic care cart within the hospital and one whiteboard in the office wrongly assigned as background. The classification indicates a few \gls{acro:FP} of furniture, such as sofas and shelves~(cp. Fig.~\ref{fig:reconstruction}). The frame processing time is faster than the mapping unless re-calculating due to graph optimization. Note: The floor and ceiling were filtered during reconstruction and manually added due to low data information with respect to the processing time.
\begin{table}
   \centering\small
    \newcolumntype{L}{p{\dimexpr.43\linewidth-2\tabcolsep-1.3333\arrayrulewidth}}   
    \caption{Perceptual performance of the initial background reconstruction for the selected environments.}
    \begin{tabularx}{1\columnwidth}{LXXX}
    \toprule
\textbf{Description} & \textbf{House} & \textbf{Hospital} & \textbf{Office}\\ 
\midrule
Processing time / s: $\mu$; $\sigma$ & 0.201; 0.113 & 0.202; 0.159 & 0.157; 0.100 \\
Latency / s:  $\mu$; $\sigma$  &  0.208; 0.153  & 0.147;  0.103 & 0.53; 0.172\\
CPU, GPU load / \%:  $\mu$; $\sigma$  & 35.9; 0 & 39.1; 0 & 21.6; 0 \\
Sum: $i$; $o$ & 58; 3129 & 103; 4453  & 124; 1566 \\ 
Precision stat./dyn. / \% & 100.0 & 98.4 & 99.2\\  
Precision semantics / \%  & 74.1 & 72.8  & 68.6 \\  
Walls: TP; FP & 37; 13   & 74; 28  & 87; 44 \\  
Tables: TP; FP  & 6; 2  & 1; 0  & 2; 1 \\  
Furniture: FP  & 15 & 14  & 13 \\  
Door, door frame: FP  & 0 & 12  & 27 \\  
Foreground instances: FP & 0 & 2 & 1\\ 
Background / m : $\mu$; $\sigma$  & 0.022; 0.059  & 0.033 0.096  & N/A \\
\bottomrule
\label{tab:backgroundsemantics}
\end{tabularx}
\end{table}

\begin{table}[tbp]
    \centering \small
    \newcolumntype{L}{p{\dimexpr.45\linewidth-2\tabcolsep-1.3333\arrayrulewidth}}   

    \caption{Perceptual long-term performance for the selected environments.}
    \begin{threeparttable}
    \begin{tabularx}{1\columnwidth}{LXXX}
    \toprule
\textbf{Description}  & \textbf{House} & \textbf{Hospital}& \textbf{Office}\\ 
\midrule
 3D detect. time / s : $\mu$; $\sigma$  & 0.047; 0.054 & 0.118; 0.122 &  0.101; 0.026 \\ 
 3D detect. latency / s :   $\mu$; $\sigma$  & \multicolumn{2}{c}{0.0774; 0.122} & 0.43;  0.232 \\
 CPU; GPU load / \% :  $\mu$   & 48.5; 12 & 51.5; 12  & 43.7; 21  \\
Classification\tnote{1} $i$: TP; FP  & 42; 3 &  297; 8 &  66; 4 \\

Sum $i$, sum $o$ & 45; 61489   &  385; 158919 & 70, 16262 \\
Loc. accuracy / m :  $\mu$; $\sigma$  & 0.014;  0.018 & 0.015; 0.012  & 0.020; 0.011 \\
People dyn. ratio  / \% & 26.8  & 18.0 & 5.6 \\
People $|v|$ / m/s : $\mu$; $\sigma$   & 0.126; 0.060 &0.157; 0.087  & 0.101; 0.191\\
People $CI$ / m/s: $c_l$; $c_u$ & [0.051; 0.288] & [0.037; 0.580] & [0.030; 0.277]\\
\bottomrule
\label{tab:data}
\end{tabularx}
 \begin{tablenotes}
      \item[1] Not present classes, such as \emph{boat} and \emph{traffic light}, have been excluded beforehand.
    \end{tablenotes}
  \end{threeparttable}
  
\label{tab:foregroundsemantics}
\end{table}

\subsubsection{Foreground Long-Term Perception} 
\begin{figure}[tb]
  \centering
  \includegraphics[width=1.0\linewidth]{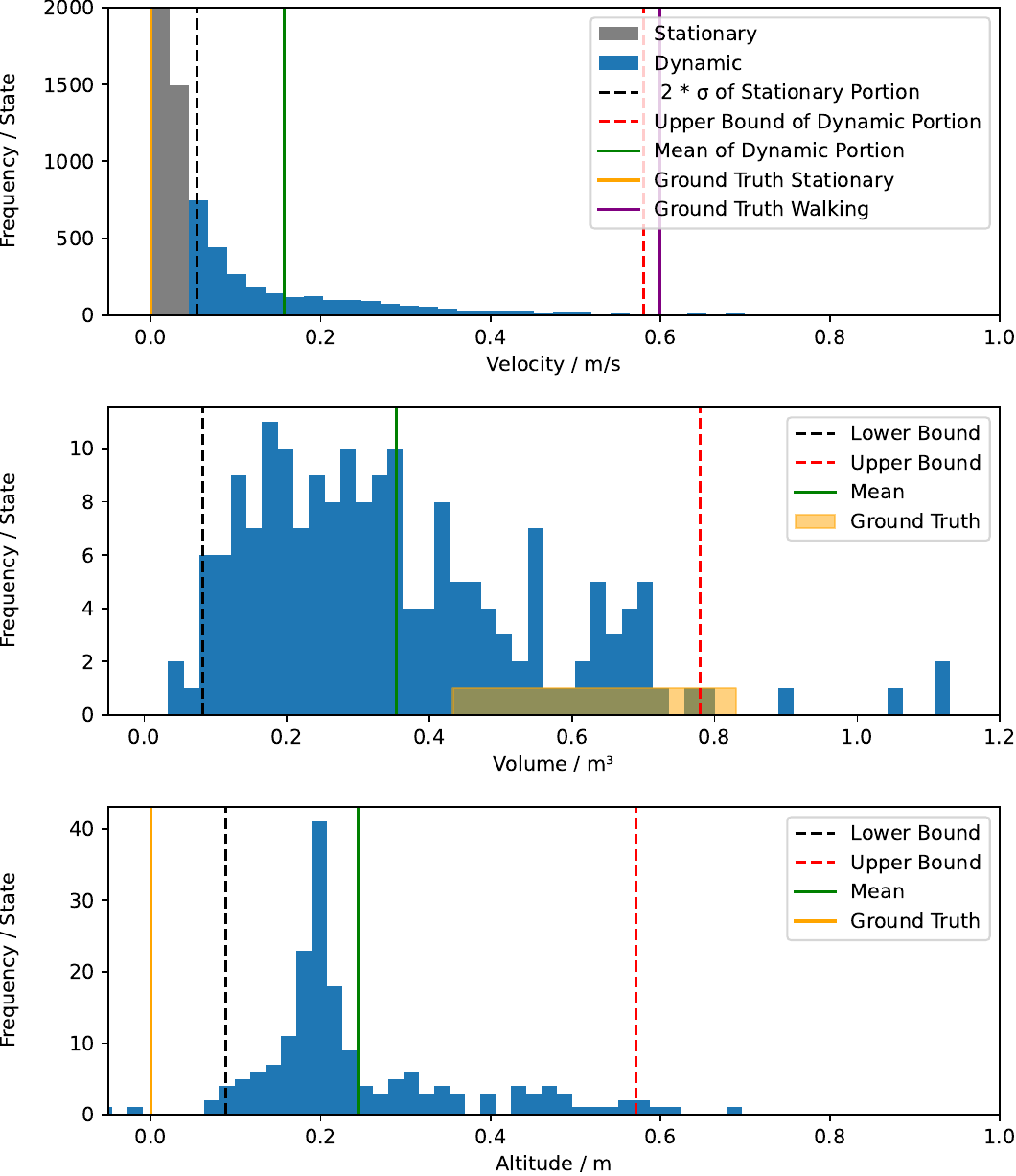}
\caption{The histograms show the distributions of determined properties of class \textit{person} in the hospital environment. The ground truth is overlayed to allow its interpretation.}
\label{fig:object_proporties}
\end{figure}

The resulting perceptual performance is summarized in Table~\ref{tab:foregroundsemantics}. Its foreground recognition pipeline achieved about 10~FPS, varying on the segmentation performance, i.e., the number of processed points after background filtering. CPU and GPU loads are comparatively low. The higher GPU load in the real-world experiments is traced back to the camera driver. A high amount of scene data with many \gls{acro:TP} instances has been acquired thanks to filtering non-present object classes. Separate experiments verify that the instance localization is precise and independent of the robot's velocity. The total number of instances includes multiple assigned ones due to a missing re-detection. The long-term analysis of the object dynamics indicates that people have been observed \SI{5.6}{\percent}~(office) to \SI{26.8}{\percent}~(house) moving. Since the simulated environments offer a known ground truth, their determined object properties are analyzed in detail. Figure~\ref{fig:object_proporties} shows them of the hospital where people systematically move from A to B with baseline dynamic portion of \SI{100}{\percent} and a constant velocity of \SI{0.6}{\meter/\second}. The results indicate that the people's velocity is \SI{97.5}{\percent} below \SI{0.58}{\meter/\second}, i.e., the prediction of the velocity is systematically too low.
The reason might be the experimental setup paired with the detection rate. The robot faces the people with a velocity difference of up to \SI{1.5}{\meter/\second}~(robot $v <=0.9m/s$). An analysis shows that due to the limited point cloud range of \SI{5}{\meter}, the robot has less than \SI{3.34}{\second} and approximately 30~detections, respectively, to determine the person's velocity. Notably, as the Kalman Filter initializes person tracks with a zero velocity, several frames must gradually accelerate the velocity and decelerate it to zero after the disappearance. Observing the object size and height shows that objects were incompletely recognized since the volume measurements are below and the altitude is higher than the ground truth. Nonetheless, the confidence interval offers a robust foundation for determining the expected range of properties. Thus, it robustifies matching and tracking for static and dynamic instances since the robot can predict known object classes' size, altitude, and dynamics.

\begin{figure}[tb]
  \centering
  \includegraphics[width=0.9\linewidth]{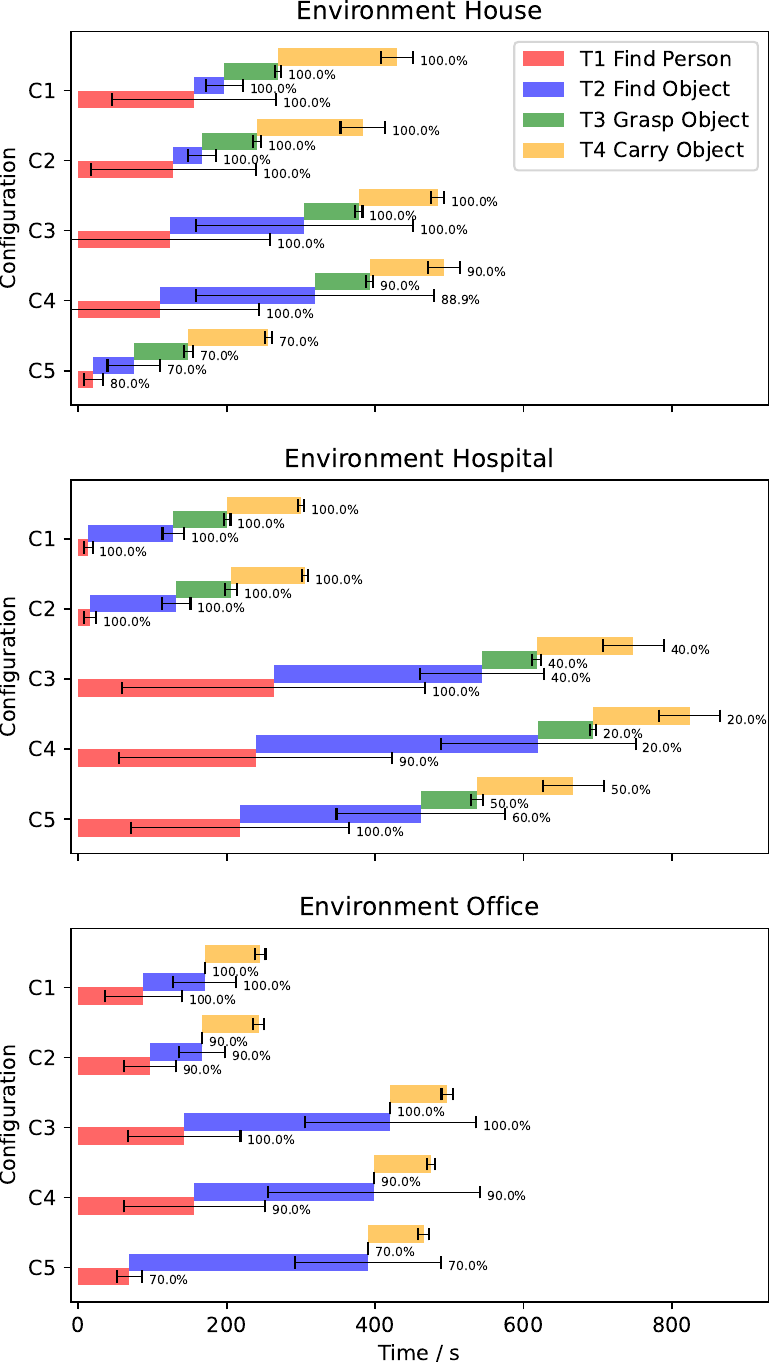}
 \caption {Task time and success rate for all ablation configurations~(C1: Full Model, C5: Basic Model) and environments. The scenario time is the aggregated task time, depicted by the mean~($n = 10$). The standard deviation per task is denoted by the black bar and the success rate on each task's right. The scenario success rate is redundant to T4 since the tasks are executed sequentially.}
\label{fig:ablation_results}
\end{figure}

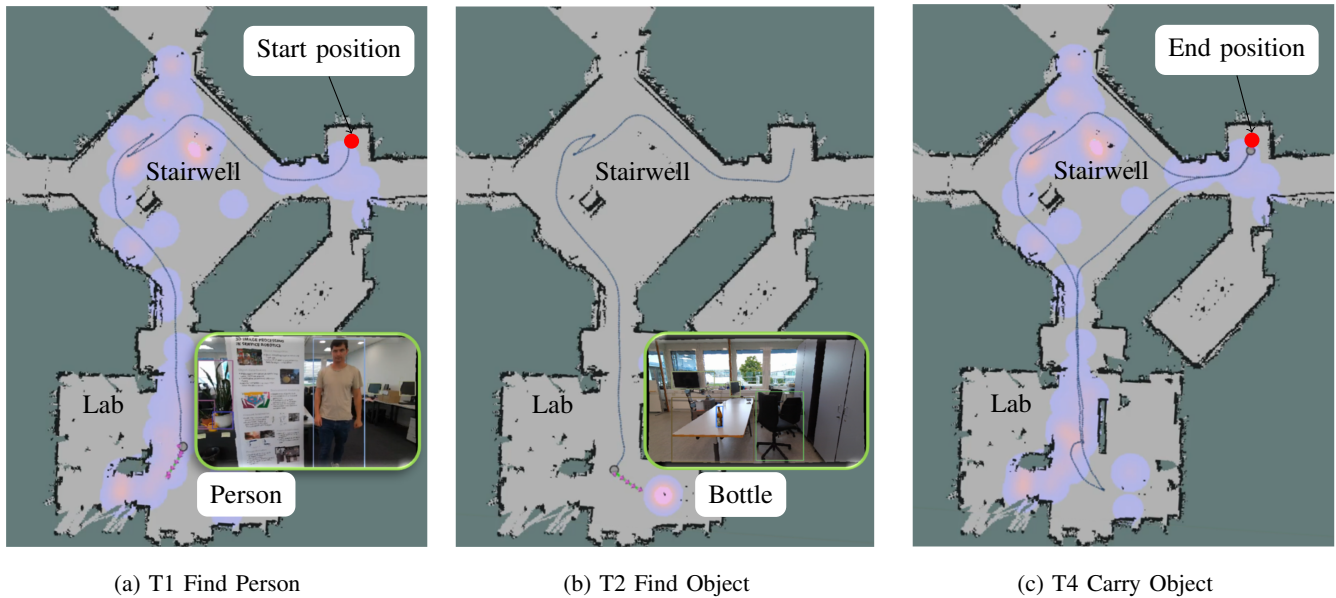
\begin{figure*}[tb]
\begin{subfigure}{.31\linewidth}
  \begin{tikzpicture}
     \centering

    \node at (0.0, 0.0) {\includegraphics[width=1.0\linewidth]{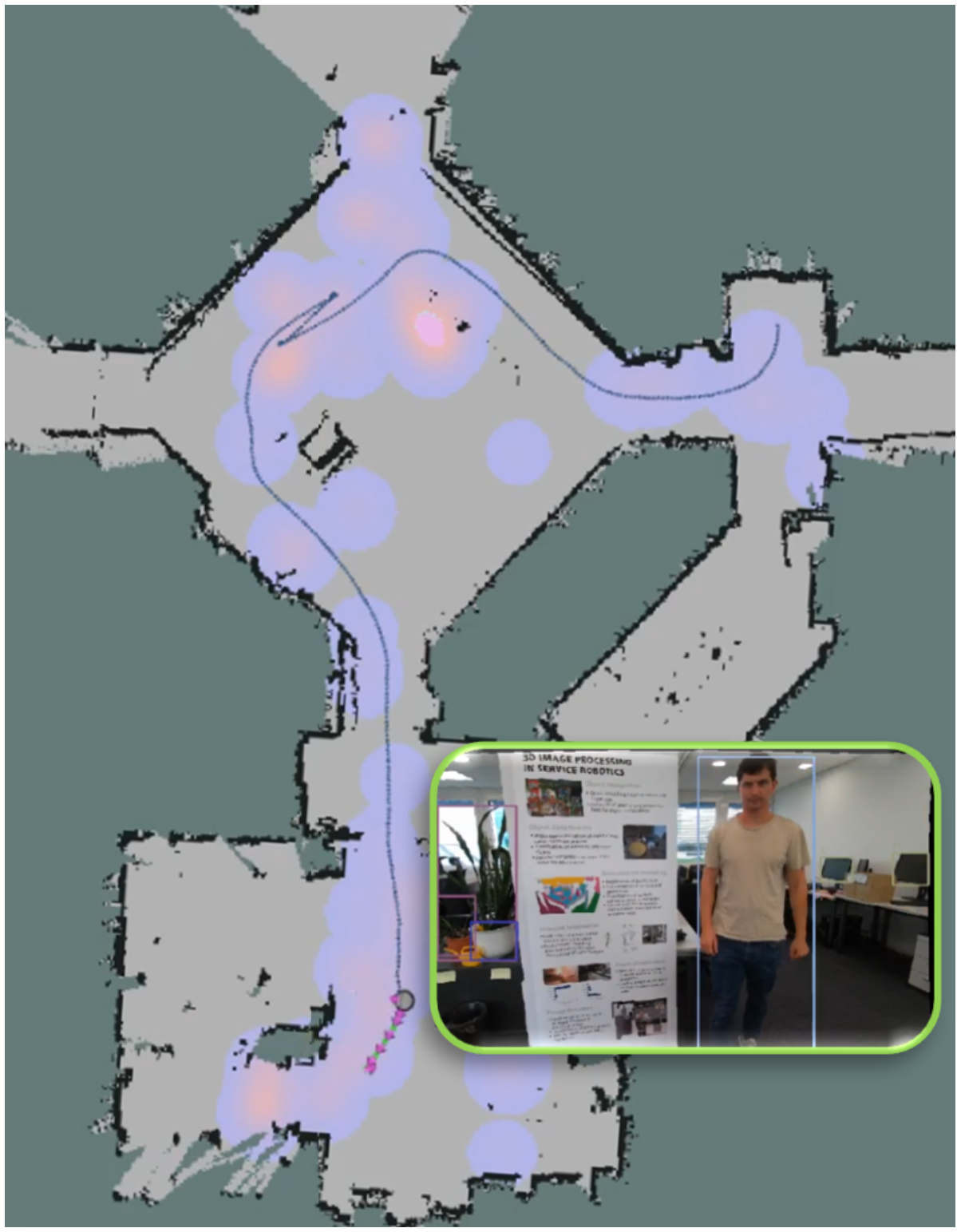}};

    \node[fill=white!30, rounded corners, inner sep=5pt](idle) at (1.5,3) {Start position};
    \node[rounded corners, inner sep=5pt](staircase) at (-0.3,1.4) {Stairwell};
    \node[rounded corners, inner sep=5pt](lab) at (-1.5,-1.7) {Lab};

    \node[fill=white!30, rounded corners, inner sep=5pt](person) at (0.4,-2.9) {Person};

    \fill[red] (1.8, 1.8) circle (0.1); 
    \draw[->] (idle.south) -- (1.8, 1.9);

    \end{tikzpicture}
    \caption{T1 Find Person}
\end{subfigure}%
\quad
\begin{subfigure}{.31\linewidth}
  \centering
    \begin{tikzpicture}
    \node at (0.0, 0.0) {\includegraphics[width=1.0\linewidth]{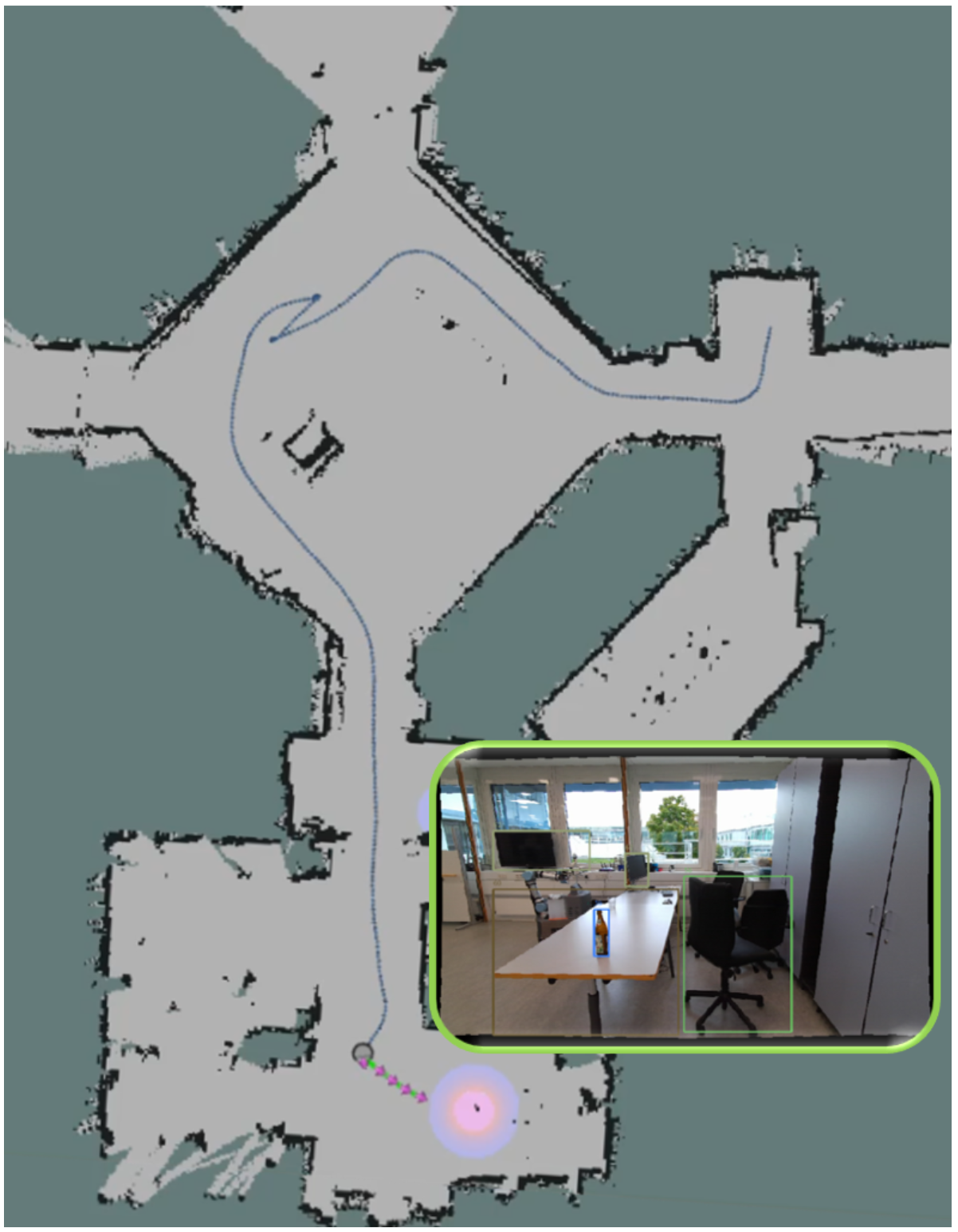}};
    \node[fill=white!30, rounded corners, inner sep=5pt](bottle) at (1,-2.9) {Bottle};
    \node[rounded corners, inner sep=5pt](staircase) at (-0.3,1.4) {Stairwell};
    \node[rounded corners, inner sep=5pt](lab) at (-1.5,-1.7) {Lab};
    
    \end{tikzpicture}
    
  \caption{T2 Find Object}
\end{subfigure}
\quad
\begin{subfigure}{.31\linewidth}
  \centering
    \begin{tikzpicture}
    \node at (0.0, 0.0) {\includegraphics[width=1.0\linewidth]{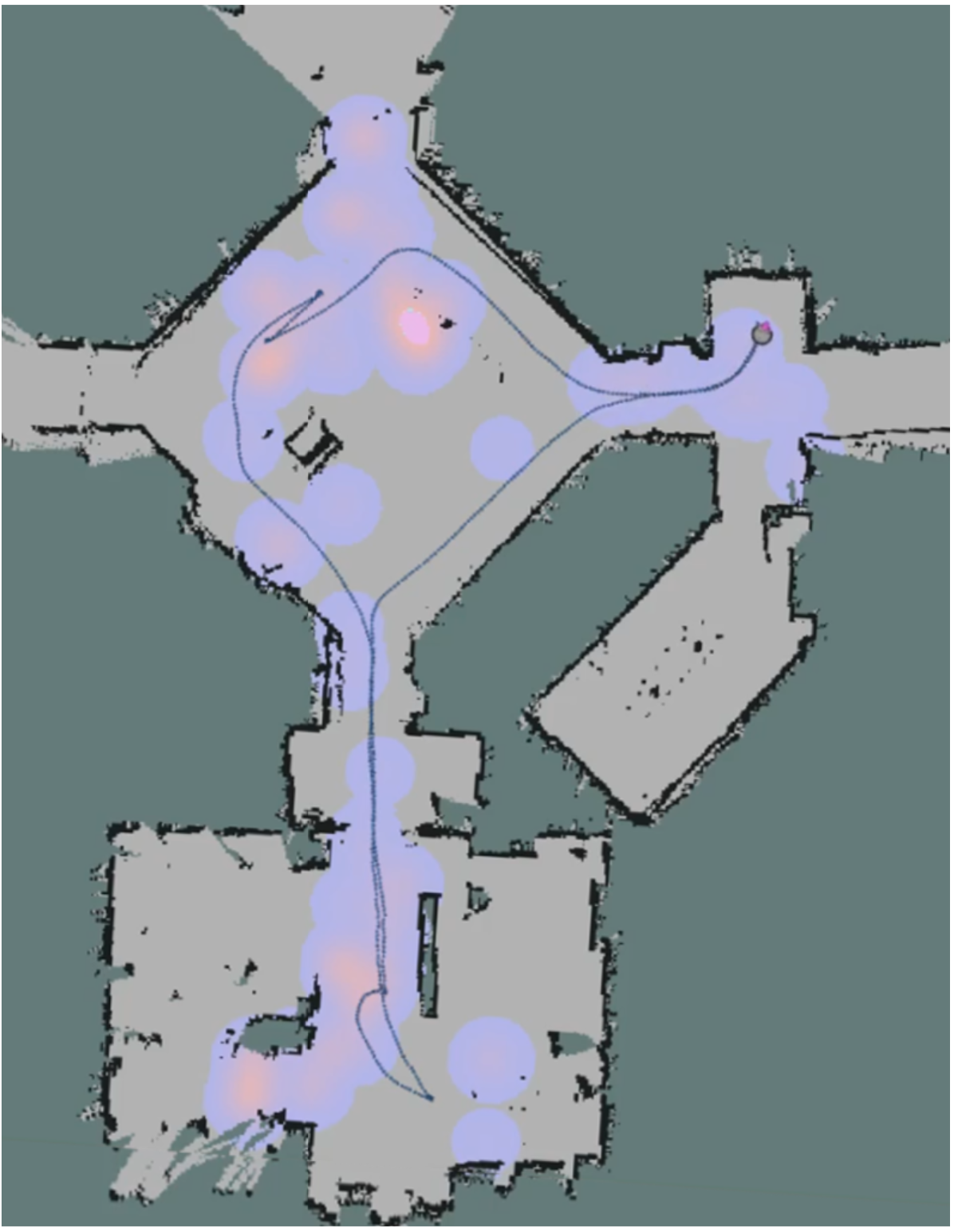}};
    \node[fill=white!30, rounded corners, inner sep=5pt](idle) at (1.5,3) {End position};
    \fill[red] (1.7, 1.8) circle (0.1); 
    \draw[->] (idle.south) -- (1.7, 1.9);
    
    \node[rounded corners, inner sep=5pt](staircase) at (-0.3,1.4) {Stairwell};
    \node[rounded corners, inner sep=5pt](lab) at (-1.5,-1.7) {Lab};

    \end{tikzpicture}
  \caption{T4 Carry Object}
\end{subfigure}
\caption{\added{Procedure of the selected tasks in the office building with the entire \gls{acro:HI}~(C1). The robot navigates across the blue trajectory using prior scene knowledge. First, the robot uses the inverse heatmap of people~(a) to search for people in heated areas, such as near the staircase and the central area of the lab. After detecting a human, the robot repeats the procedure with the inverse heatmap of bottles~(b) to find the closest one. Finally, the robot returns to the idle position, where the heatmap of people optimizes its navigation to avoid crowded areas~(c).}}

\label{fig:office_tasks}
\end{figure*}
\subsubsection{Ablation Study} 
The ablation study removes system components to obtain insights into their functionality, a widely applied approach in fields like machine learning. It starts with the entire \gls{acro:HI} and then systematically ablates fragments as described by the configurations in Table~\ref{tab:ablationconfig} until a so-called \emph{basic model} is released. The basic model only comprises the \gls{acro:SLAM} and the object detector,  adopted from the early approach presented in~\cite{Odabasi2022}. The resulting time and the success rate of all configurations and tasks are compared in Fig.~\ref{fig:ablation_results}. A task is considered unsuccessful if either a failure occurs or its execution exceeds \SI{600}{\second}.

\subsubsection*{C1: Entire Model}
\begin{figure}[tb]
  \centering
  \includegraphics[width=0.95\linewidth]{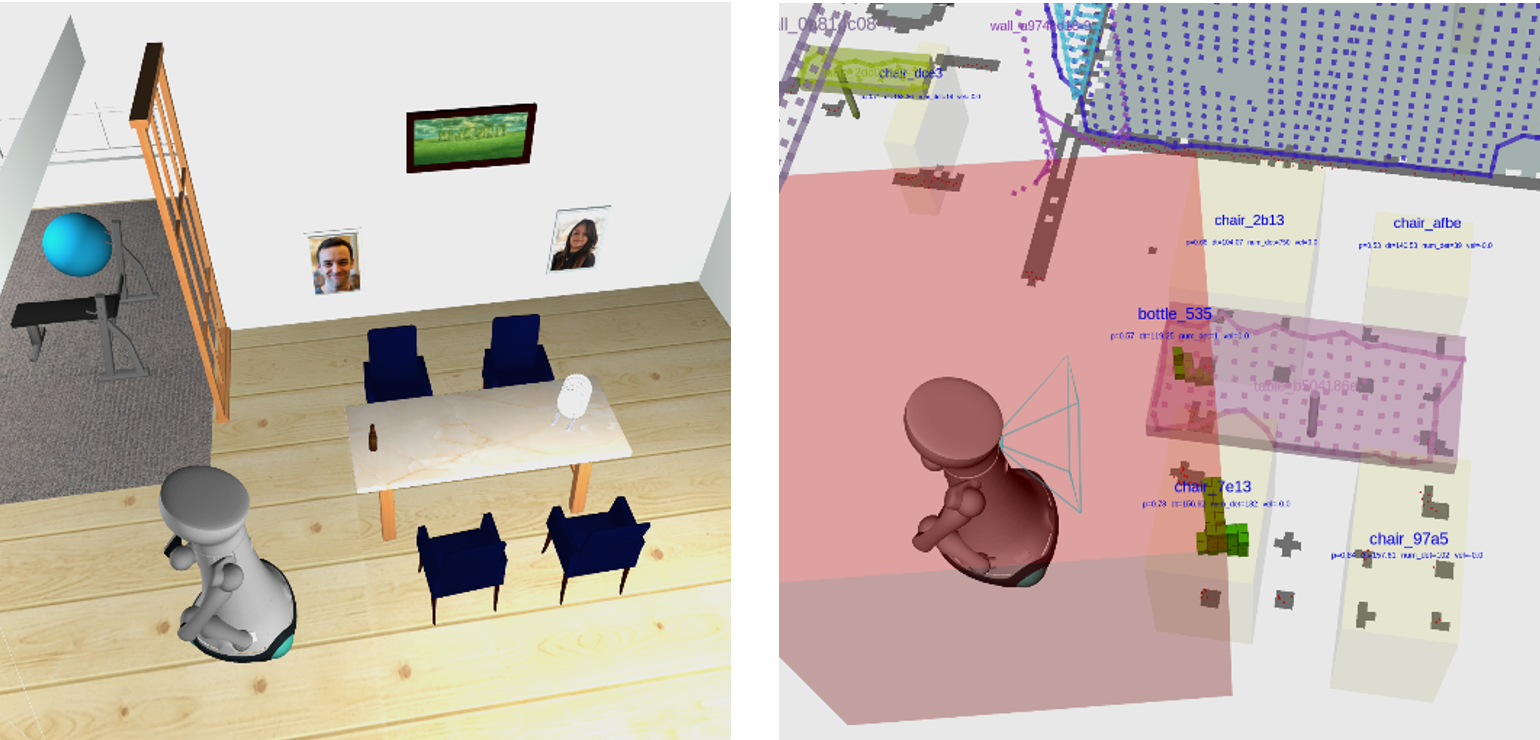}
  \caption{\added{Third person view on the manipulation task.} The manipulation interface provides the static background, represented by basic shapes, and the dynamic foreground, represented by an occupancy grid with semantics and $ODS$, within the manipulation workspace (red cube).}
  \label{figures:manipulation}
\end{figure}

\added{An exemplary procedure within the office building is visualized in Fig.~\ref{fig:office_tasks}.} The robot executes the tasks using its acquired scene knowledge starting at its idle position as highlighted in Fig.~\ref{fig:envs} and Fig.~\ref{fig:reconstruction}. In T1, the robot employs an inverse heatmap generated for class "people" to navigate into prior crowded areas to search for a person. The task time variability is higher in environments with random people movement, such as when comparing the house and the hospital. In T2, the robot can constrain its search for the bottle due to its knowledge of the environment-specific altitude and the usage of its heatmap. T3, executed in simulation only, involves the robot approaching recognized bottles using a method described in a prior study~\cite{Graf2019}, followed by the bottle grasping. \added{The grasping uses the 3D bounding box center of the recognized rotation-symmetric bottle, as described in our previous study}~\cite{Odabasi2022}. The semantically enhanced regions~(cp. Fig.~\ref{figures:manipulation}) \added{around the bottle provide a high-level understanding of the local workspace.} They allow for optimized obstacle handling due to \added{the knowledge of static background areas, represented by primitive shapes, and foreground regions, represented by a voxel grid assigned with the $ODS$}. In T4, the robot utilizes similar to T1 the heatmaps of people to avoid crowded areas for fast and safe navigation. 

\subsubsection*{C2: Feature Learning Removed}

\begin{figure}[tb]
\centering
\begin{subfigure}{.42\linewidth}\centering
    \includegraphics[width=1.0\textwidth]{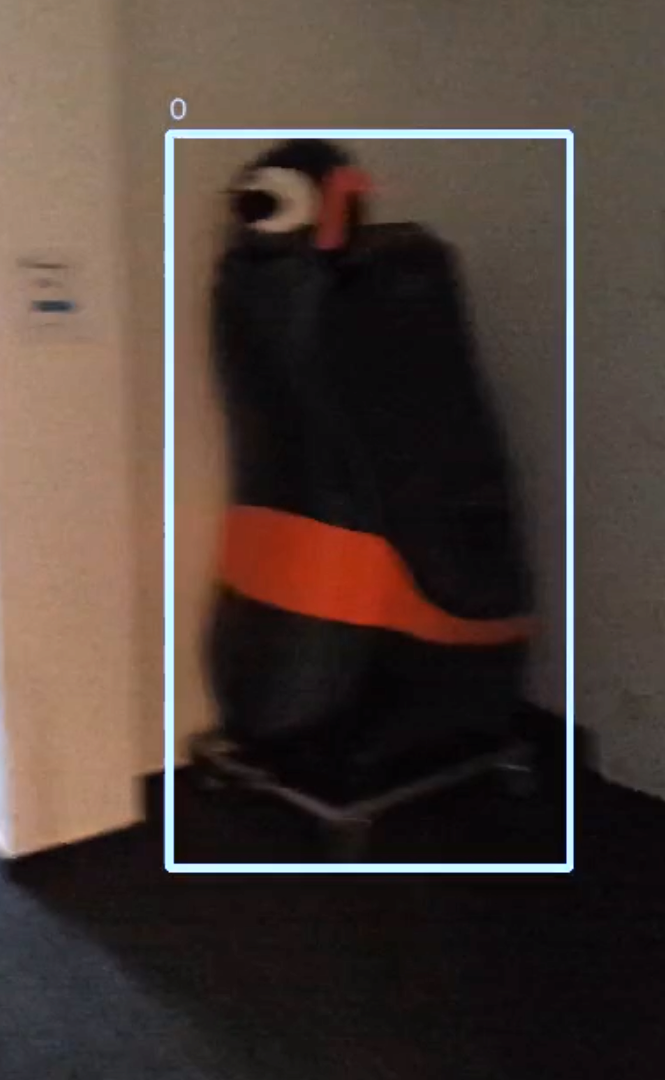}
    \caption {Robot detected as person}
\label{fig:fp_robot}
\end{subfigure}
\hfill
\begin{subfigure}{.55\linewidth}\centering
    \includegraphics[width=1.0\textwidth]{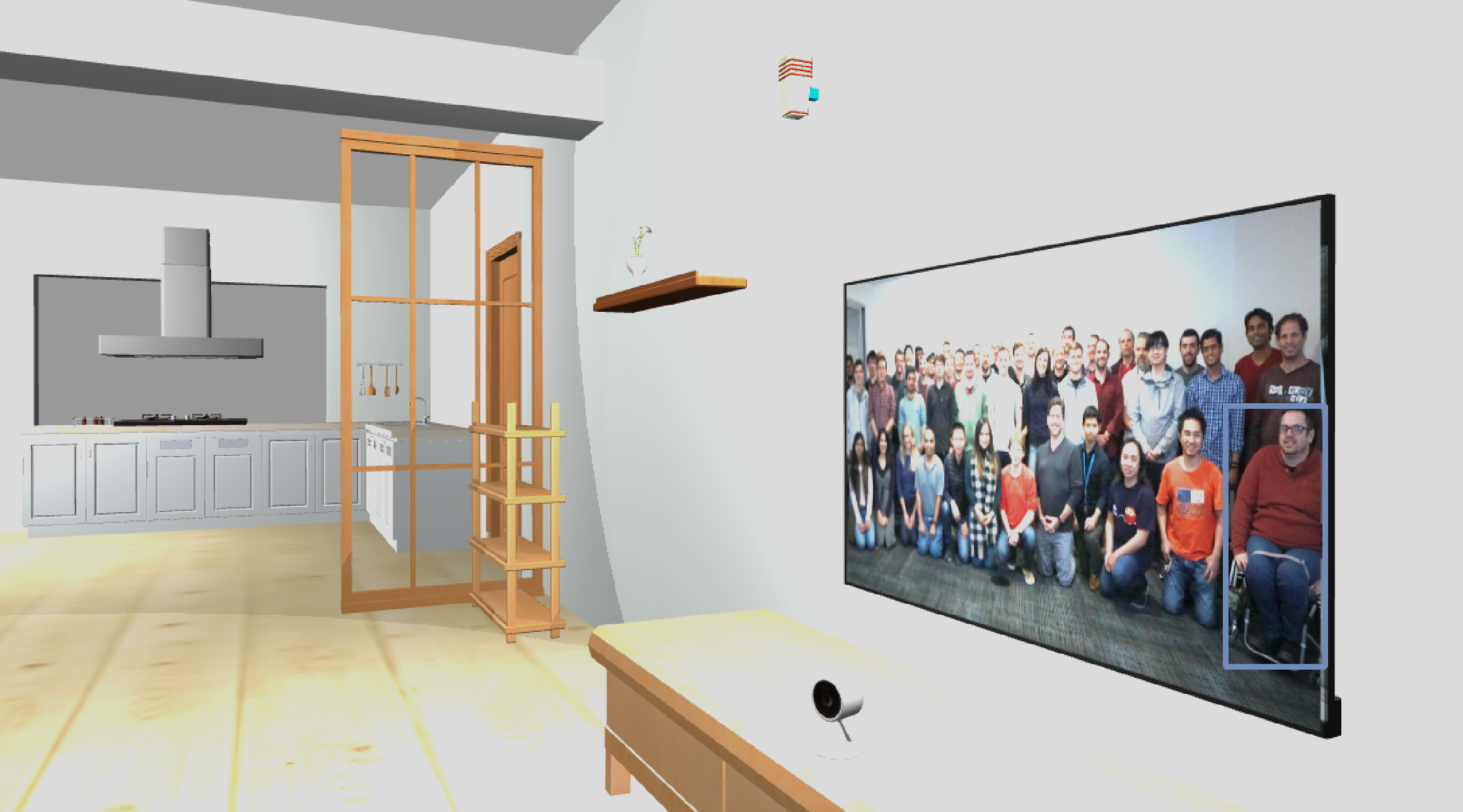}
    \caption {Recognized people on a TV}
    \label{fig:fp_people}
    \includegraphics[width=1.0\textwidth]{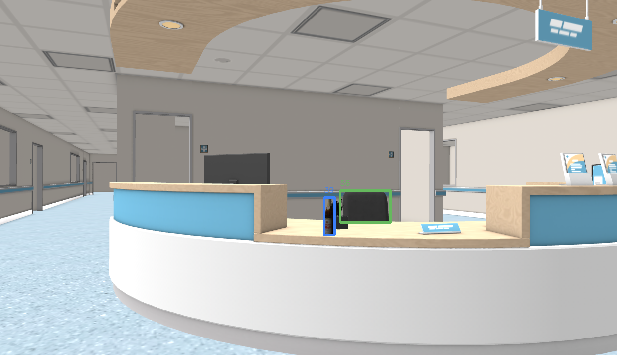}
    \caption {Bottle on a cluttered desk}
\label{fig:desk}
\end{subfigure}%
\caption{Challenges in object recognition experiments include: (a) Occasional misidentification of another robot as a person in the office. (b) Intermittent failure of C5 to recognize people on the TV in the house during T1. (c) Difficulty in detecting small bottles in the hospital due to size, occlusions, and cluttered background, leading to task timeouts.}
\label{fig:issues}
\end{figure}

Excluding the component of feature learning implies that the robot lacks environment-specific object knowledge regarding the typical altitude, size, and dynamics. The results indicate that T1 and T2 do not yield significant changes. However, one FP appears in the office where another robot has been identified as person~(cp.~Fig.~\ref{fig:fp_robot}), a hard example that could have occurred with the entire approach as well due to its similar size. 
In T2, the robot adapts to no longer assuming a specific altitude for bottles, relying on heatmaps to limit potential bottle areas with consistent results. T3 involves approaching and requesting scene data for manipulation, but the robot loses awareness of potential dynamic elements, necessitating the maximization of safety distances. While impacting safety, this change does not significantly affect task time and success rates. Lastly, in T4, the robot navigates without calculating object-class-related safety distances but achieves consistent results by avoiding frequent and crowded areas, as only people are considered dynamic. Concluding, removing the perceptual learning component does not significantly impact the task time but slightly increases the error rate.

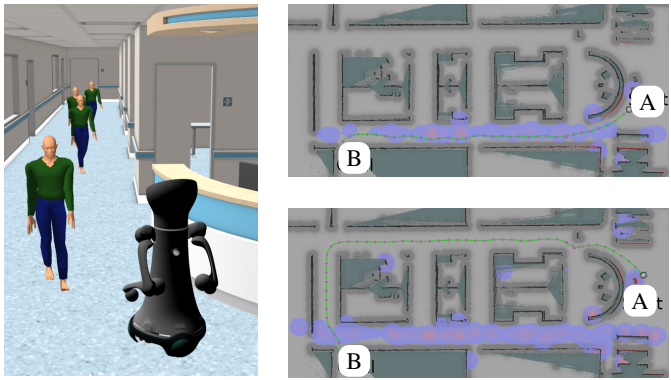
\begin{figure}[tb]
\begin{subfigure}{.38\linewidth}
\centering
    \begin{tikzpicture}
    \node at (0.0, 0) {\includegraphics[width = 1.0\linewidth]{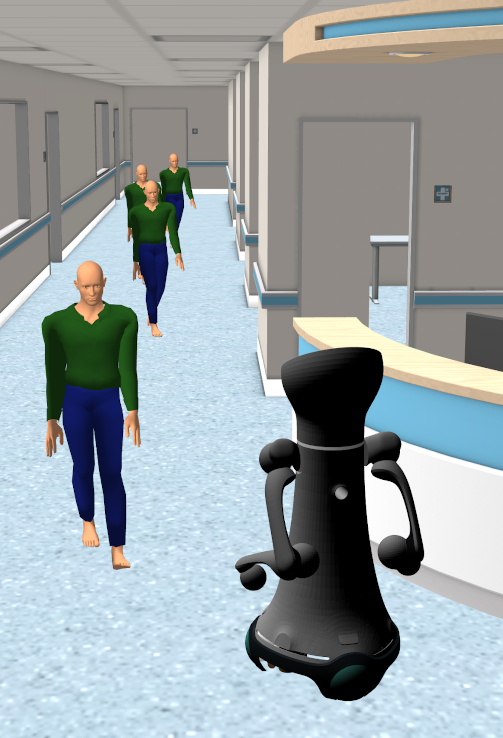}};
    \end{tikzpicture}
\caption{Crowded corridor with walking people challenges the robot navigation.}

\label{fig:hospital_corridor}
\end{subfigure}
\hfill
\begin{subfigure}{.57\linewidth}
\centering
    \begin{tikzpicture}
    \scriptsize
    \node at (0.0, 1.35) {\includegraphics[width = 1.0\linewidth]{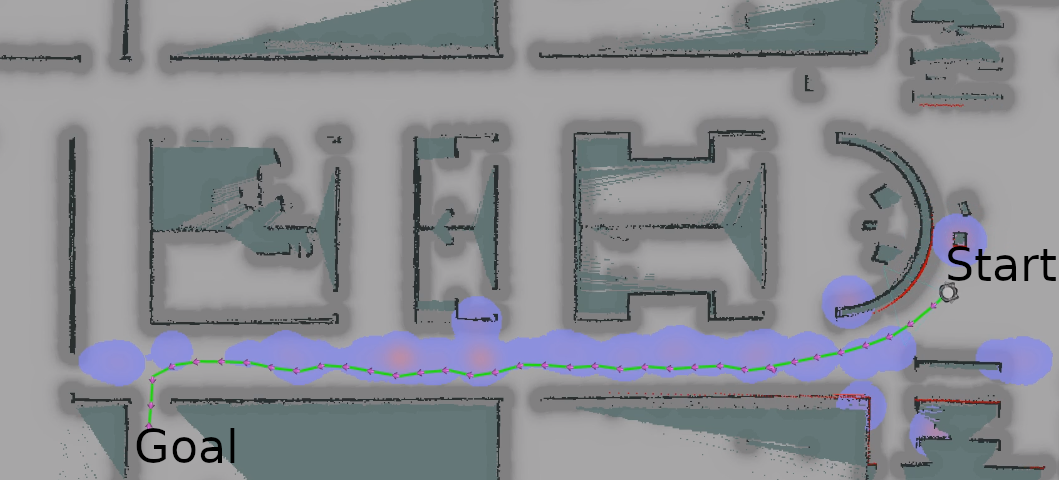}};
    
    \node at (0.0, -1.35) {\includegraphics[width = 1.0\linewidth]{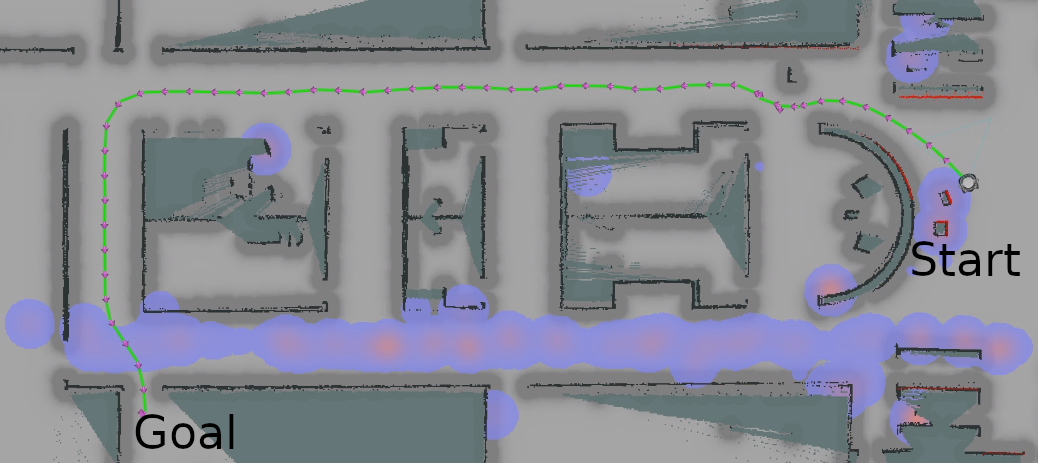}};
    
    \node[fill=white!30, rounded corners, inner sep=3pt](a) at (2.25,1.2) {\small A};
    \node[fill=white!30, rounded corners, inner sep=3pt](b) at (-1.65,0.45) { \small B};

      \node[fill=white!30, rounded corners, inner sep=3pt](a) at (2.15,-1.45) {\small A};
    \node[inner sep=3pt](dummy) at (-1.66,-2.35) { \small '};

    \node[fill=white!30, rounded corners, inner sep=3pt](b) at (-1.65,-2.25) { \small B};

    \end{tikzpicture}
  \caption{Map after one~(top) and two~(bottom) hours. The heatmap~(colored) is overlayed with the costmap~(grayscale).}
  \label{fig:hospital_heat}
\end{subfigure}
\caption{Heatmaps generated from prior scene knowledge increase the path costs in the crowded southern hospital corridor~(a).  After collecting sufficient scene information, the robot path (green) switches to the empty northern corridor~(b).
}
\label{fig:heatmaps_hospital}
\end{figure}


\subsubsection*{C3: Heatmaps Removed}

This configuration removes the heatmaps to ignore historical object knowledge for behavior control but keeps the knowledge base since it is necessary for the MOT. The removal changes in T1 that the robot has to perform a coverage of the entire area due to missing prior information. 
The impact without heatmaps varies across environments, \SI{5}{\percent} faster in the house but a substantial \SI{1535}{\percent} slower in the hospital. In the unconstrained office building the task time of T1 has been increased by \SI{47}{\percent} as a result of the longer path. T2 presents a similar challenge when searching for a bottle. The task is finished \SI{398}{\percent} slower in the house, \SI{175}{\percent} slower in the hospital, and \SI{275}{\percent} slower in the office. Moreover, the success rate decreases, such as in the hospital, where 6/10 runs have been aborted as they exceed the time limit. Although the bottle is reached in time, bottle detection is challenging due to its size, and the cluttered scene, which causes occlusions~(cp. Fig.~\ref{fig:desk}). T3 indicates that ablating heatmaps and the object aging feature have no effect. In T4, the robot is no longer avoiding crowded areas and experiences slower task times, especially in the hospital, with a significant \SI{27}{\percent} increase since the robot takes the shorter path through the crowded corridor instead of the empty one as before~(cp. Fig.~\ref{fig:heatmaps_hospital}). The change in task time, though insignificant in other environments, underscores the relevance of productivity and safety during navigation.

\subsubsection*{C4: Basic Model with Foreground Recognition} 
Removing background recognition means that foreground recognition must process the entire point cloud. This leads to a significant increase in frame processing time and latency for foreground recognition in all tasks. For example, in a hospital setting, the latency increases from an average of $\mu$ = \SI{0.077}{\second}, $\sigma$ = \SI{0.122}{\second}~(C1-C3) to $\mu = $\SI{0.130}{\second}, $\sigma$ = \SI{0.47}{\second}~(C4), representing an approximately \SI{69}{\percent} higher processing delay. The absence of background recognition alters task behavior in the following ways:
In T1, the absence of background recognition leads to a reduced success rate since people in portraits on walls and on a TV~(Fig.~\ref{fig:fp_people}) have been wrongly identified. T2 performs similarly to before but with a lower bottle detection accuracy due to the occasional merging of the desk with bottle points to a unified segment. In T3, the missing background makes the distinction between static and dynamic elements unfeasible. This leads to increased data processing and higher latency, with a \SI{70}{\percent} average increase in dynamic obstacles and a \SI{69}{\percent} increase in latency. The manipulation planning process is also affected, requiring consideration of more data points for obstacle-aware motion planning. T4 indicates that the absence of background information amplifies cluster sizes for the manipulation task, and a slower \gls{acro:MOT} pipeline compromises the robot's reactivity and safety.

\subsubsection*{C5: Basic Model}
This configuration comprises the SLAM and object detector, similar to an earlier approach outlined in~\cite{Odabasi2022}. In T1, the robot achieves faster detection due to stand-alone usage of the 2D detector without range limitations but experiences a low success rate due to FP on portraits and TV. T2 has been executed similarly to before, but also profiting from the longer range for detecting the bottle. The bounding boxes of the \gls{acro:2D} object detector are projected into \gls{acro:3D} space to localize objects for T3. T3 faces complexities in object identification for grasping, lacking table support, and the ability to perceive clusters. While manipulation can access static background data, the foreground presents unprocessed raw data without clustering or semantic labels. Finally, T4 was performed as before.
\section{Discussion}
\label{sec:discuss}


Related integrated approaches lack a common baseline, and benchmarks \added{evaluating the perception based on A-B motions} are not designed for long-term applications. The introduced benchmark addresses this issue by presenting feasible everyday tasks for today's mobile robots.  The experiments showcase a cross-environment deployment in which the split between dynamic foreground and static plane-based background achieved \SI{>98}{\percent} precision and accelerate the foreground recognition by \SI{70}{\percent}, thus agreeing with related work that it can be generalized as a valid approach for indoor environments. In contrast to related approaches, it solely requires the exchange of the reused 2D detector, being available for many domains. Related work uses fully integrated, thus indivisible approaches, e.g., with DL-based semantic segmentation, which is problematic for computation and represents a challenge to find a proper dataset.

The ablation study with its results in Fig.~\ref{fig:ablation_results} proves that the \gls{acro:HI} empowers mobile robots to perform typical tasks faster and more reliably \added{as it reduces typical errors.} Knowledge interpretation allows the adaptation to the environment in operation, as demonstrated with feature learning and heatmaps. Although feature learning achieved only minor improvements, it offers potential beyond the experiment setup, e.g., for socially aware navigation by using object tracks with the $ODS$ to define an appropriate safety distance. Similarly, the features help the manipulation for motion planning since the $ODS$ provides the experienced reactivity, thus, the safety distance can be automatically maximized for dynamic objects and minimized for static objects. The experiments showcase that if the objects of interest are situated in distinct areas, heatmaps accelerate the scenario by up to \SI{61}{\percent}, but experience a slight deceleration when object classes are equally distributed. However, an unconstrained real-world experiment demonstrated a \SI{47}{\percent} acceleration. One unexpected effect occurs in C5, where the search tasks have been executed faster than before in the house. This effect \added{is traced back to the setup of relying on pure 2D detections, which have no} range limitation. Thus, we propose \added{using 2D detections as an initial region proposal to improve the search strategy of C1-C4.} 

In total, 550 tasks have been executed. As the tasks are repeatedly executed in a constant setting, the impact of each perception component has been revealed. While the benchmark allows quantifying the performance of the \gls{acro:HI}, its comparison to related approaches is not yet applicable due to experimental effort. Thus, we want to encourage other researchers to take over our benchmark. \added{Nonetheless, when comparing the basic model (C5) the performance of the 3rd party components, i.e., RTAB-Map and YOLO, is well known. To measure the impact of \gls{acro:HI}-specific components is impossible with an unconstrained setup, i.e., it would require sticking back to an A-B motion dataset. Noteworthy, as no other component than YOLO sticks on the GPU, the background filtering reduces the data points, and interpretation components are triggered only ad-hoc, the computational efficiency is not yet limiting the application as underlined by a preliminary version of the \gls{acro:HI} running on low-cost hardware}~\cite{Graf2022b}. \added{This example also proves an easy adaptation to other object classes and applications.}

\section{Conclusion}
\label{sec:conclude}


The bioinspired approach of taking over concepts of human perception represents a novel approach to holistic scene perception for mobile robots. Our prior study~\cite{Graf2022a} highlights that holistic scene perception approaches, i.e., which cover the entire process from recognition to interpretation of acquired knowledge, are rare, particularly because interpretation techniques are rarely discovered in robotic research.
This paper introduces \gls{acro:HI}, a technical solution that solves the seamless integration of existing and novel methods to generate a holistic scene perception by imitating two major neuroscience concepts. First, the triplet split of human perception allowing to define functionalities and their purpose clearly. The recognition generates scene observations and converts them to aggregated instances, a multi-layer knowledge base represents these entities, and spatio-temporal analyses interpret the obtained scene knowledge. Second, the preattentive and postattentive split, a sub-concept of the recognition process that splits the foreground and background.

Simulation and real-world experiments validate that the approach significantly improves the performance of typical tasks in the open-world context.  The experiments prove that the holistic approach empowers mobile service robots to execute popular tasks faster and more reliably than stand-alone SLAM and object detection solutions since open-world problems are considered. Although a gap to human capabilities remains, the \gls{acro:HI} is a significant step toward multifunctional robots as its concepts and experiments, which obtained a high amount of data (over 0.5 million observations for a few hours), highlight the potential for future work. 
The background classification lacks performance which could be improved with a more complex approach to add environment-specific knowledge, such as furniture, doors, and corridors. The selected object detector achieved proper results but might require a wider knowledge beyond the classes of the COCO dataset. Thus, we emphasize \added{merging concepts of open-vocabulary mapping to improve managing open-world problems.  We see high potential in connecting the knowledge base to a Large Language Model to exploit the full potential of the data.} Finally, we propose to deepen perceptual learning, such as for the re-identification of objects, to handle reappearance and corner cases.

\section{Acknowledgment}
This research has received funding from the German Ministry of Economic Affairs, Labour and Tourism Baden-Württemberg by the AI Innovation Center \textit{Learning Systems and Cognitive Robotics} under grant agreement No 017-180036~(841022).

\bibliographystyle{IEEEtran}
\bibliography{main}

\newpage




\begin{IEEEbiography}[{\includegraphics[width=1in,height=1.25in,clip,keepaspectratio]{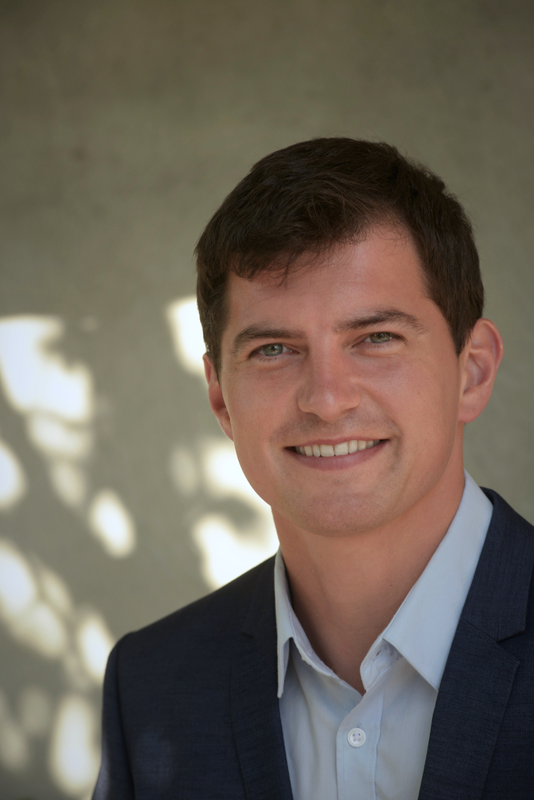}}]{Florenz Graf}
received his M.S. in Mechanical Engineering from Leibniz University Hannover, Germany. Since 2017, he has been working on scene perception as a research associate within the Department of Robot and Assistive Systems at Fraunhofer IPA in Stuttgart, Germany. He is currently pursuing his Ph.D. degree at the University of Stuttgart, Germany. His research interests include scene perception and human-robot interaction for mobile robots in open-world settings.
\end{IEEEbiography}

\begin{IEEEbiography}[{\includegraphics[width=1in,height=1.25in,clip,keepaspectratio]{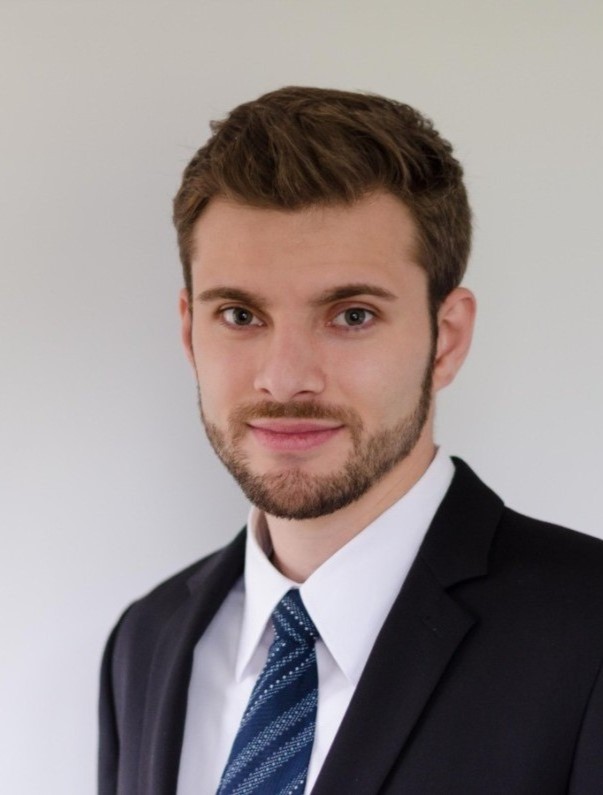}}]{Jochen Lindermayr} is a researcher at Fraunhofer IPA. After graduating as M.Sc. Electrical Engineering and Information Technology at University of Stuttgart in 2017, he joined the Department Robot and Assistive Systems at Fraunhofer IPA in Stuttgart, Germany. He focuses on computer vision and machine learning for mobile service robot applications. Currently, he is pursuing his Ph.D. degree at the University of Stuttgart, Germany.

\end{IEEEbiography}

\begin{IEEEbiography}[{\includegraphics[width=1in,height=1.25in,clip,keepaspectratio]{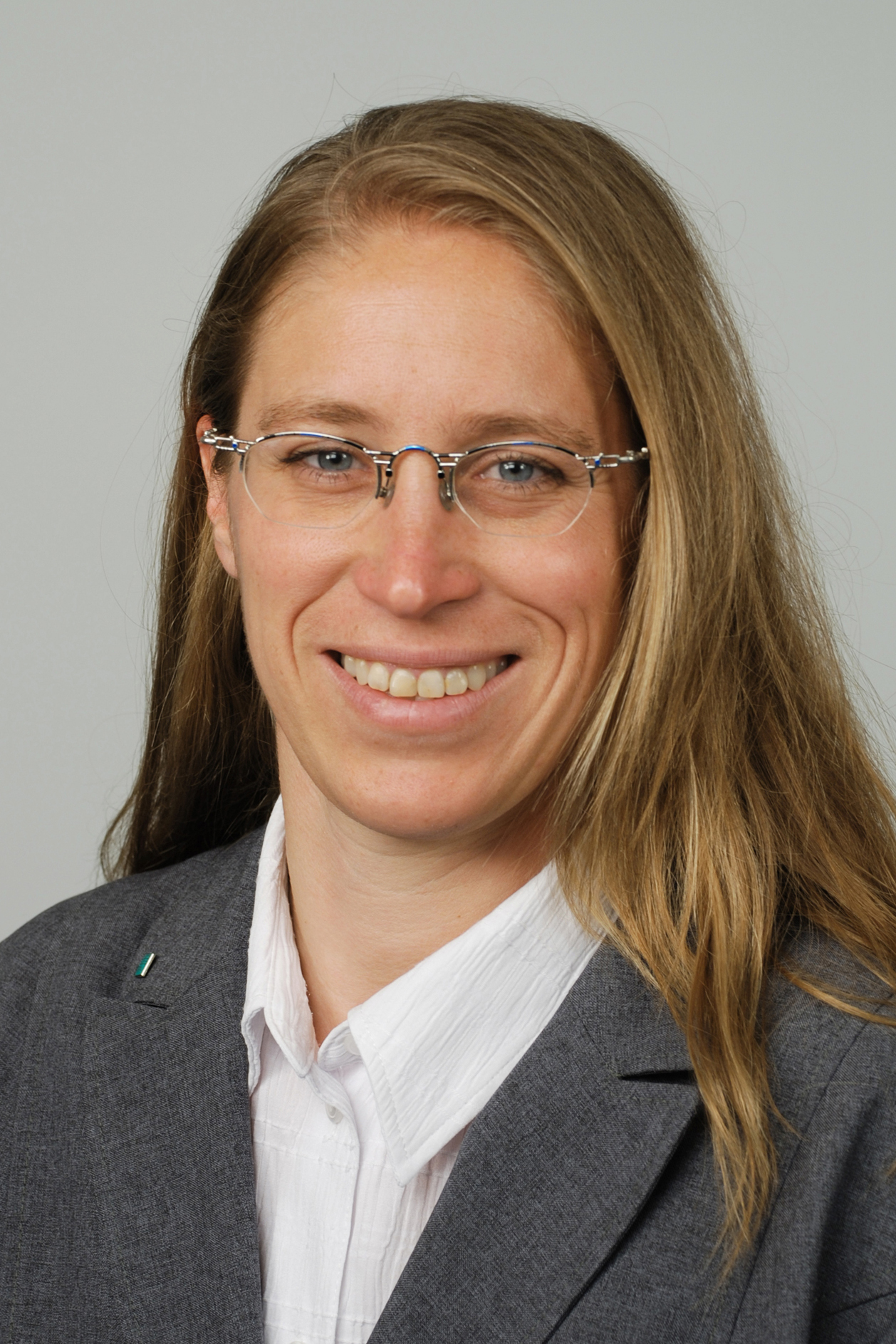}}]{Birgit Graf}
received her diploma in Computer Science from Stuttgart University, Germany in 1999 and completed her PhD on “An Adaptive Guidance System for Robotic Walking Aids” in 2008. She is manager of the Assistive Robotics Group at Fraunhofer IPA, which develops robotic solutions for health care applications. Her research interests include the analysis of user needs and automation potentials in residential care facilities or hospitals and the design of robotic solutions suited to improve working conditions of the staff and quality of care.
\end{IEEEbiography}

\begin{IEEEbiography}[{\includegraphics[width=1in,height=1.25in,clip,keepaspectratio]{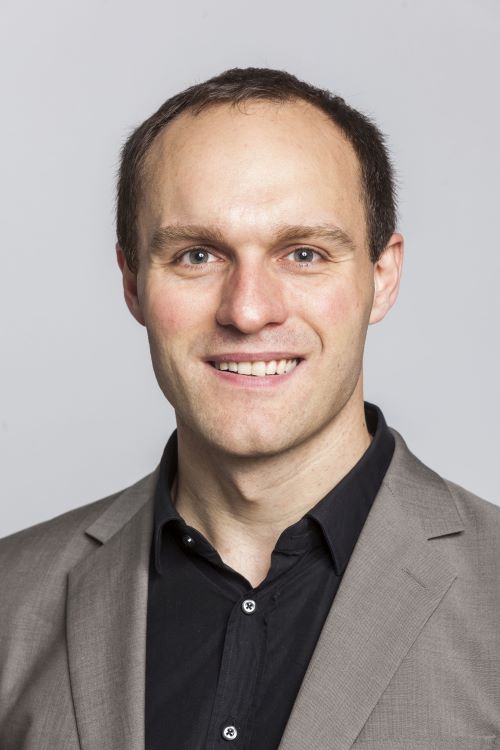}}]{Werner Kraus} received his degree as Diplom-Ingenieur in mechanical engineering from the Karlsruhe Institute of Technology in 2011. Since then, he is with Fraunhofer IPA in different positions and received his PhD on force-control of cable-driven parallel robots in 2015 from the University of Stuttgart. Since June 2019, he is head of department robot and assistive systems. His research interests include: industrial and service robots, cognitive robotics, control and applications of cable-driven parallel robots.
\end{IEEEbiography}

\begin{IEEEbiography}[{\includegraphics[width=1in,height=1.25in,clip,keepaspectratio]{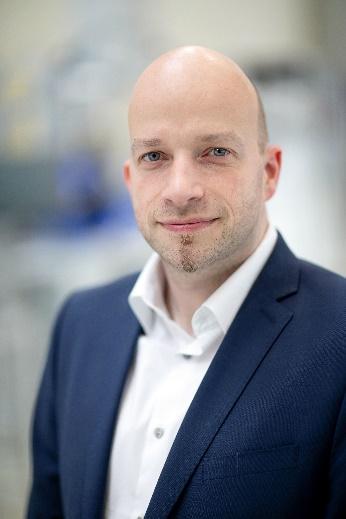}}]{Marco Huber}
received his diploma, Ph.D., and habilitation degrees in computer science from the Karlsruhe Institute of Technology (KIT), Germany, in 2006, 2009, and 2015, respectively. From June 2009 to May 2011, he was leading a research group at the Fraunhofer IOSB, Karlsruhe, Germany. He then took on various management positions in the industry until September 2018. Since October 2018 he is full professor with the University of Stuttgart. At the same time, he is director of the Center for Cyber Cognitive Intelligence (CCI) and of the Department for Image and Signal Processing with Fraunhofer IPA in Stuttgart, Germany. His research interests include machine learning, planning and decision-making, image processing, and robotics.
\end{IEEEbiography}

\end{document}